\documentclass[10pt,twocolumn,letterpaper]{article}
\pdfoutput=1
\usepackage{cvpr}
\usepackage{times}
\usepackage{epsfig}
\usepackage{graphicx}
\usepackage{amsmath}
\usepackage{amssymb}

\usepackage{amsmath,amsfonts}

\newcommand{\X}{\mathcal{X}}

\newcommand{\Z}{\mathcal{Z}}

\newcommand{\R}{\mathbb{R}}

\newcommand{\U}{\mathcal{U}}
\newcommand{\mP}{\mathcal{P}}

\newcommand{\E}{\mathbb{E}}

\newcommand{\bQ}{{\boldsymbol{Q}}}

\newcommand{\bD}{{\boldsymbol{D}}}
\newcommand{\bC}{{\boldsymbol{C}}}

\newtheorem{theorem}{Theorem}
\newtheorem{lemma}{Lemma}

\newcommand\numberthis{\addtocounter{equation}{1}\tag{\theequation}}
\newcommand\independent{\protect\mathpalette{\protect\independenT}{\perp}}
\def\independenT#1#2{\mathrel{\rlap{$#1#2$}\mkern2mu{#1#2}}}
\usepackage{algorithm}
\usepackage[noend]{algpseudocode}
\usepackage{multicol, multirow}
\usepackage{booktabs}
\usepackage{mathtools}
\usepackage{enumitem}
\setitemize{noitemsep,topsep=0pt,parsep=0pt,partopsep=0pt}
\usepackage{lipsum}
\usepackage{placeins}
\usepackage[breaklinks=true,bookmarks=false]{hyperref}

\cvprfinalcopy 

\def\cvprPaperID{****} 

\begin{document}

\title{Optimal Transport Classifier: Defending Against Adversarial Attacks by Regularized Deep Embedding}
\author{
Yao Li$^{2}$, Martin Renqiang Min$^1$, Wenchao Yu$^{3}$, Cho-Jui Hsieh$^3$, Thomas C. M. Lee$^2$, and Erik Kruus$^1$\\
$^1$NEC Laboratories America\\
$^2$University of California, Davis\\
$^3$University of California, Los Angeles
}
\maketitle

\begin{abstract}
Recent studies have demonstrated the vulnerability of deep convolutional neural networks against adversarial examples. Inspired by the observation that the intrinsic dimension of image data is much smaller than its pixel space dimension and the vulnerability of neural networks grows with the input dimension, we propose to embed high-dimensional input images into a low-dimensional space to perform classification. However, arbitrarily projecting the input images to a low-dimensional space without regularization will not improve the robustness of deep neural networks. Leveraging optimal transport theory, we propose a new framework, Optimal Transport Classifier (OT-Classifier), and derive an objective that minimizes the discrepancy between the distribution of the true label and the distribution of the OT-Classifier output. Experimental results on several benchmark datasets show that, our proposed framework achieves state-of-the-art performance against strong adversarial attack methods.

\end{abstract}
\vspace{-12pt}
\section{Introduction}
Deep neural networks (DNNs) have been widely used for tackling numerous machine learning problems that were once believed to be challenging. With their remarkable ability of fitting training data, DNNs have 
achieved revolutionary successes in many fields such as computer vision, natural language progressing, and robotics. However, they were shown to be vulnerable to adversarial examples that are generated by adding carefully crafted perturbations to original images. The adversarial perturbations can arbitrarily change the network's prediction but often too small to affect human recognition~\cite{szegedy2013intriguing,kurakin2016adversarial}. 
This phenomenon brings out security concerns for practical applications of deep learning.


Two main types of attack settings have been considered in recent research~\cite{goodfellow6572explaining,carlini2017adversarial,chen2017zoo,papernot2017practical}: black-box and white-box settings. In the black-box setting, the attacker can provide any inputs and receive the corresponding predictions. 
However, the attacker cannot get access to the gradients or model parameters under this setting; whereas in the white-box setting, the attacker is allowed to analytically compute the model's gradients, and have full access to the model architecture and weights. In this paper, we focus on defending against the white-box attack which is the harder task.


\begin{figure}
    \centering
    \includegraphics[width=0.47\textwidth]{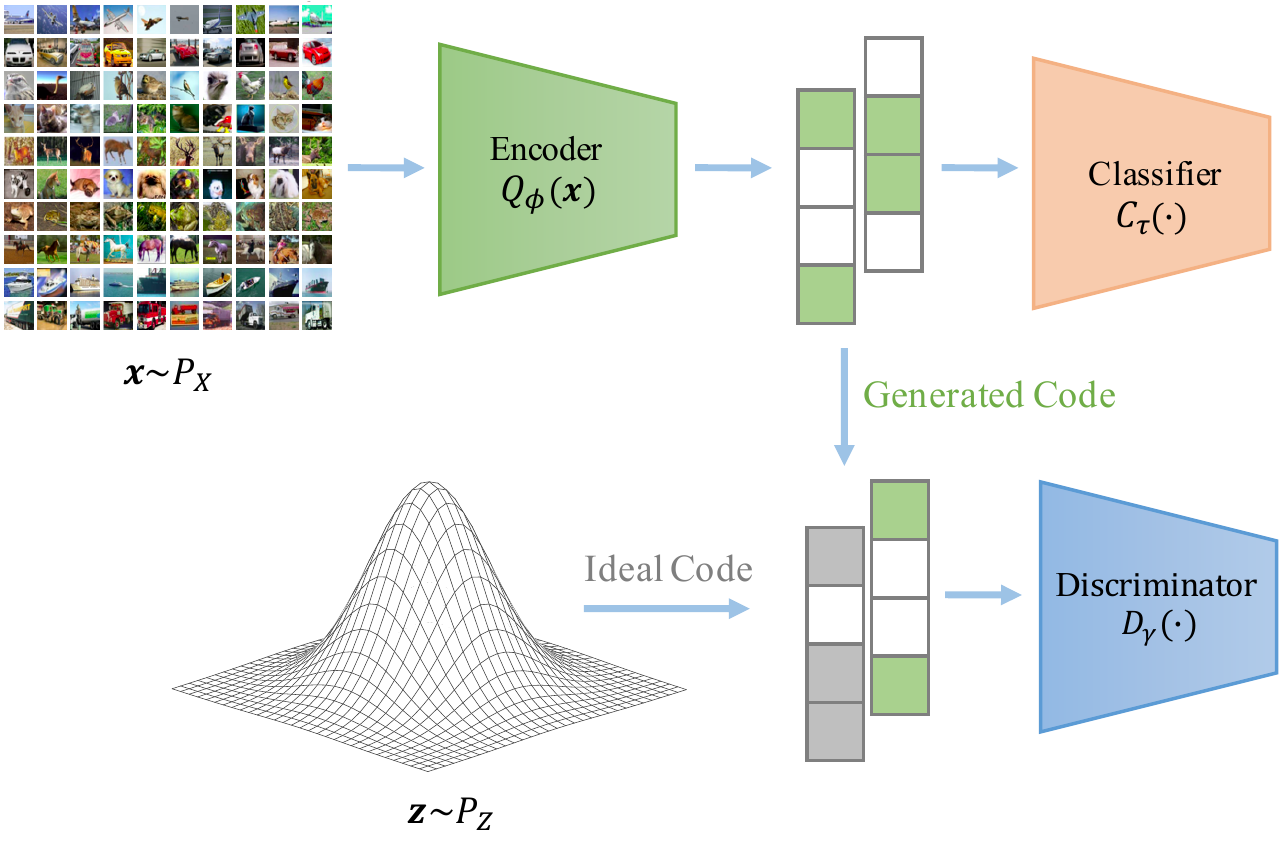}
    \caption{Overview of OT-Classifier framework}
    \label{fig:d_cla}
\vspace{-10pt}
\end{figure}

Recent work~\cite{simon2018adversarial} presented both theoretical arguments and an empirical one-to-one relationship between input dimension and adversarial vulnerability, showing that the vulnerability of neural networks grows with the input dimension. Therefore, reducing the data dimension may help improve the robustness of deep neural networks. Furthermore, a consensus in the high-dimensional data analysis community is that, a method working well on the high-dimensional data is because the data is not really of high-dimension~\cite{levina2005maximum}. These high-dimensional data, such as images, are actually embedded in a much lower dimensional space. Hence, carefully reducing the input dimension may improve the robustness of the model without sacrificing performance. 

Inspired by the observation that the intrinsic dimension of image data is actually much smaller than its pixel space dimension~\cite{levina2005maximum} and the vulnerability of a model grows with its input dimension~\cite{simon2018adversarial}, we propose a defense framework that embeds input images into a low-dimensional space using a deep encoder and performs classification based on the latent embedding with a classifier network.
However, arbitrarily projecting input images to a low-dimensional space based on a deep encoder does not guarantee improving the robustness of the model, because there are a lot of mapping functions including pathological ones from the raw input space to the low-dimensional space capable of minimizing the classification loss. To constrain the mapping function, we employ distribution regularization in the embedding space leveraging optimal transport theory. We call our new classification framework  Optimal Transport Classifier (OT-Classifier). To be more specific, we introduce a discriminator in the latent space which tries to separate the generated code vectors from the encoder network and the ideal code vectors sampled from a prior distribution, i.e., a standard Gaussian distribution. Employing a similar powerful competitive mechanism as demonstrated by Generative Adversarial Networks~\cite{goodfellow2014generative}, the discriminator enforces the embedding space of the model to follow the prior distribution. 

In our OT-Classifier framework, the encoder and discriminator structures together project the input data to a low-dimensional space with a nice shape, then the classifier performs prediction based on the low-dimensional embedding. Based on the optimal transport theory, the proposed OT-Classifier minimizes the discrepancy between the distribution of the true label and the distribution of the framework output, thus only retaining important features for classification in the embedding space.  With a small embedding dimension, the effect of the adversarial perturbation is largely diminished through the projection process.

We compare OT-Classifier with other state-of-the-art defense methods on MNIST, CIFAR10, STL10 and Tiny Imagenet. Experimental results demonstrate that our proposed OT-Classifier outperforms other defense methods by a large margin. To sum up, this paper makes the following three main contributions: 
\begin{itemize}[noitemsep,topsep=0pt,parsep=0pt,partopsep=0pt]
\item A novel unified end-to-end robust deep neural network framework against adversarial attacks is proposed, where the input image is first projected to a low-dimensional space and then classified.
\item An objective is induced to minimize the optimal transport cost between the true class distribution and the framework output distribution, guiding the encoder and discriminator to project the input image to a low-dimensional space without losing important features for classification. 
\item Extensive experiments demonstrate the robustness of our proposed OT-Classifier framework under the white-box attacks, and show that OT-Classifier combined with adversarial training outperforms other state-of-the-art approaches on several benchmark image datasets.
\end{itemize}

\vspace{-5pt}
\section{Related Work}
In this section, we summarize related work into three categories: attack methods, defense mechanisms and optimal transport theory. We first discuss different white-box attack methods, followed by a description of different defense mechanisms against these attacks, and finally optimal transport theory.

\subsection{Attack Methods}

Under the white-box setting, attackers have all information about the targeted neural network, including network structure and gradients. Most white-box attacks generate adversarial examples based on the gradient of loss function with respect to the input. An algorithm called fast gradient sign method (FGSM) was proposed in \cite{goodfellow6572explaining} which generates adversarial examples based on the sign of gradient. 
Many other white-box attack methods have been proposed recently~\cite{moosavi2016deepfool,chen2017ead,madry2017towards,carlini2017towards}, and among them C\&W and PGD attacks have been widely used to test the robustness of machine learning models. 

\textbf{C$\&$W attack: }The adversarial attack method proposed by Carlini and Wagner~\cite{carlini2017towards} is one of the strongest white-box attack methods. They formulate the adversarial example generating process as an optimization problem. The proposed objective function aims at increasing the probability of the target class and minimizing the distance between the adversarial example and the original input image. Therefore, C$\&$W attack can be viewed as a gradient-descent based adversarial attack.

\textbf{PGD attack: } The projected gradient descent attack is proposed by~\cite{madry2017towards}, which finds adversarial examples in an $\epsilon$-ball of the image. The PGD attack updates in the direction that decreases the probability of the original class most, then projects the result back to the $\epsilon$-ball of the input. An advantage of PGD attack over C$\&$W attack is that it allows direct control of distortion level by changing $\epsilon$, while for C$\&$W attack, one can only do so indirectly via hyper-parameter tuning. 

Both C$\&$W attack and PGD attack have been frequently used to benchmark the defense algorithms due to their effectiveness~\cite{athalye2018obfuscated}. In this paper, we mainly use $l_\infty$-PGD untargeted attack to evaluate the effectiveness of the defense method under white-box setting.

 Instead of crafting different adversarial perturbation for different input image, an algorithm was proposed by \cite{moosavi2017universal} to construct a universal perturbation that causes natural images to be misclassified. However, since this universal perturbation is image-agnostic, it is usually larger than the image-specific perturbation generated by PGD and C\&W.

\subsection{Defense Mechanisms}

Many works have been done to improve the robustness of deep neural networks. To defend against adversarial examples, defenses that aim to increase model robustness fall into three main categories: i) augmenting the training data with adversarial examples to enhance the existing classifiers~\cite{madry2017towards,na2017cascade,goodfellow6572explaining}; ii) leveraging model-specific strategies to enforce model properties such as smoothness~\cite{papernot2016distillation}; and, iii) trying to remove adversarial perturbations from the inputs~\cite{xie2017mitigating,samangouei2018defense,meng2017magnet}. We select three representative methods that are effective under white-box setting. 

\textbf{Adversarial training:} Augmenting the training data with adversarial examples can increase the robustness of the deep neural network. Madry et al.~\cite{madry2017towards} recently introduced a min-max formulation against adversarial attacks. The proposed model is not only trained on the original dataset but also adversarial example in the $\epsilon$-ball of each input image.

\textbf{Random Self-Ensemble:} Another effective defense method under white-box setting is RSE~\cite{liu2017towards}. The authors proposed a ``noise layer'', which fuses output of each layer with Gaussian noise. They empirically show that the noise layer can help improve the robustness of deep neural networks. The noise layer is applied in both training and testing phases, so the prediction accuracy will not be largely affected.

\textbf{Defense-GAN:} 
Defense-GAN~\cite{samangouei2018defense} leverages the expressive capability of GANs to defend deep neural networks against adversarial examples. It is trained to project input images onto the range of the GAN's generator to remove the effect of the adversarial perturbation.  Another defense method that uses the generative model to filter out noise is MagNet proposed by~\cite{meng2017magnet}. However, the differences between OT-Classifier and the two methods are obvious. 
OT-Classifier focus on reducing the dimension, and performing classification based on the low-dimensional embedding, while Defense-GAN and MagNet mainly apply the generative model to filter out the adversarial noise, and both Defense-GAN and MagNet perform classification on the original dimension space. \cite{samangouei2018defense} showed that Defense-GAN is more robust than MagNet, so we only compare with Defense-GAN in the experiment.


\subsection{Optimal Transport Theory}

There are various ways to define the distance or divergence between the target distribution and the model distribution. In this paper, we turn to the optimal transport theory\footnote{More details available at  \url{https://optimaltransport.github.io/slides/}}, which provides a much weaker topology than many others. In real applications, data is usually embedded in a space of a much lower dimension, such as a non-linear manifold. \textit{Kullback-Leibler} divergence, \textit{Jensen-Shannon} divergence and \textit{Total Variation} distance are not sensible cost functions when learning distributions supported by lower dimensional manifolds~\cite{arjovsky2017wasserstein}. In contrast, the optimal transport cost is more sensible in this setting. 

Kantorovich's distance induced by the optimal transport problem is given by
\begin{align*}
    W_c(P_Y, P_\bC) \coloneqq \inf_{\Gamma\in\mP(Y\sim P_Y, U\sim P_\bC)}\E_{(Y,U)\sim\Gamma}\left\{c(Y,U)\right\},
\end{align*}
where $\Gamma\in\mP(Y\sim P_Y, U\sim P_\bC)$ is the set of all joint distributions of $(Y,U)$ with marginals $P_Y$ and $P_\bC$, and $c(y,u):\U\times\U\mapsto\R_{+}$ is any measurable cost function. $W_c(P_Y, P_\bC)$ measures the divergence between probability distributions $P_Y$ and $P_\bC$. 

When the probability measures are on a metric space, the $p$-th root of $W_c$ is called the $p$-Wasserstein distance. Recently, Tolstikhin~\cite{tolstikhin2017wasserstein} introduced a new algorithm to build a generative model of the target data distribution based on the Wasserstein distance. The proposed generative model can generate samples of better quality, as measured by the FID score.

\vspace{-5pt}
\section{Proposed Framework: Optimal Transport Classifier}
We propose a novel defense framework, OT-Classifier, which aims at projecting the image data to a low-dimensional space to remove noise and stabilize the classification model by minimizing the optimal transport cost between the true label distribution $P_Y$ and the distribution of the OT-Classifier output ($P_\bC$). 
The encoder and discriminator structures together help diminish the effect of the adversarial perturbation by projecting input data to a space of lower dimension, then the classifier part performs classification based on the low-dimensional embedding. 

\subsection{Notations}
In this paper, we use $l_\infty$ and $l_2$ distortion metrics to measure similarity. We report $l_\infty$ distance in the normalized $[0,1]$ space, so that a distortion of $0.031$ corresponds to $8/256$, and $l_2$ distance as the total root-mean-square distortion normalized by the total number of pixels. 

We use calligraphic letters for sets (i.e., $\X$), capital letters for random variables (i.e., $X$), and lower case letters for their values (i.e., $x$). The probability distributions are denoted with capital letters (i.e., $P_X$) and corresponding densities with lower case letters (i.e., $p_X$). 

Images $X\in \X= \mathbb{R}^d$ are projected to a low-dimensional embedding vector $Z\in \Z = \mathbb{R}^k$ through the encoder $\bQ_\phi$. The discriminator $\bD_\gamma$ discriminates between the generated code $\tilde{Z}\sim \bQ_\phi(Z|X)$ and the ideal code $Z\sim P_Z$. The classifier $\bC_\tau$ performs classification based on the generated code $\tilde{Z}$, producing output $U\in \U=\mathbb{R}^m$, where $m$ is the number of classes. The label of $X$ is denoted as $Y \in \U$. An overview of the framework is shown in Figure~\ref{fig:d_cla}.

\subsection{Framework Details}
At training stage, the encoder $\bQ_\phi$ first maps the input $x$ to a low-dimensional space, resulting in generated code ($\tilde{z}$). Another ideal code ($z$) is sampled from the prior distribution, and the discriminator $\bD_\gamma$ discriminates between the ideal code (positive data) and the generated code (negative data). The classifier ($\bC_\tau$) predicts the image label based on the generated code ($\tilde{z}$). Details of training process can be found in Algorithm~\ref{alg:d_cla}.


\begin{algorithm}
\caption{Training OT-Classifier}\label{alg:d_cla}
\begin{algorithmic}[1]
\State \textbf{Input:} Regularization coefficient $\lambda>0$, encoder $\bQ_\phi$, discriminator $\bD_\gamma$, and classifier $\bC_\tau$.
\State \textbf{Note:} $\ell$ stands for the cross-entropy loss.
 \While{$(\phi, \gamma, \tau)$ not converged}
 \State Sample $\{(x_1,y_1),...,
 (x_n,y_n)\}$ from the training set
 \State Sample $\{z_1,...,z_n\}$ from the prior $P_Z$
 \State Sample $\tilde{z}_i$ from $\bQ_\phi(Z|x_i)$ for $i=1,...,n$
 \State Update $\bD_\gamma$ by ascending the following objective by 1-step Adam:
 \begin{align*}
     \frac{\lambda}{n}\sum_{i=1}^n \bD_\gamma(z_i)- \bD_\gamma(\tilde{z}_i)
 \end{align*}
 \State Update $\bQ_\phi$ and $\bC_\tau$ by descending the following objective by 1-step Adam:
 \begin{align*}
     \frac{1}{n}\sum_{i=1}^n\ell(\bC_\tau(\bQ_\phi(x_i)),y_i)
 \end{align*}
 \State Update $\bQ_\phi$ by ascending the following objective by 1-step Adam:
 \begin{align*}
     \frac{\lambda}{n}\sum_{i=1}^n\bD_\gamma(\bQ_\phi(x_i))
 \end{align*}
\EndWhile
\State \textbf{end while}

\end{algorithmic}
\end{algorithm}

At inference time, only the encoder $\bQ_\phi$ and the classifier $\bC_\tau$ are used. The input image $x$ is first mapped to a low-dimensional space by the encoder ($\tilde{z}=\bQ_\phi(x)$), then the latent code $\tilde{z}$ is fed into the classifier to obtain the predicted label.

Our framework can be combined with other state-of-the-art defense methods, such as adversarial training. Since the dimension of the input images are reduced to a much lower dimension, adversarial training also benefits from this dimension reduction. In the experiments, we combine OT-classifier with adversarial training and compare it with other defense methods. 

\subsection{Theoretical Analysis}
The OT-Classifier framework embeds important classification features 
by minimizing the discrepancy between the distribution of the true label ($P_Y$) and the distribution of the framework output ($P_\bC$). In the framework, the classifier ($P_\bC(U|Z)$) maps a latent code $Z$ sampled from a fixed distribution in a latent space $\Z$, to the output $U\in \U=\mathbb{R}^m$. The density of OT-Classifier output is defined as follow: 
\begin{align}
    p_\bC(u) \coloneqq \int_\Z p_\bC(u|z)p_Z(z)dz, \ \ \forall u\in \U.
\label{eq:p_c}
\end{align}


In this paper we apply standard Gaussian as our prior distribution $P_Z$, but other priors may be used for different cases. Assume there is an oracle $f:\X\mapsto\U$ assigning the image data ($X\in\X$) its true label ($Y\in \U$). To minimize the optimal transport cost between the distribution of the true label ($P_Y$) and the distribution of the OT-Classifier output ($P_\bC$), it is sufficient to find a conditional distribution $\bQ(Z|X)$ such that its marginal distribution $\bQ_Z$ is identical to the prior distribution $P_Z$.

\begin{theorem} For $P_\bC$ as defined above with a deterministic $P_\bC(U|Z)$ and any function $\bC:\Z\mapsto\U$
\begin{align*}
   &\inf_{\Gamma\in\mP(Y\sim P_Y, U\sim P_\bC)}\E_{(Y,U)\sim\Gamma}\left\{\ell(Y,U)\right\}\\
    &=\inf_{\bQ:\bQ_Z=P_Z}\E_{P_X}\E_{\bQ(Z|X)}\left\{\ell(f(X),\bC(Z))\right\},
\end{align*}
where $\Gamma\in\mP(Y\sim P_Y, U\sim P_\bC)$ is the set of all joint distributions of $(Y,U)$ with marginals $P_Y$ and $P_\bC$, and $\ell(y,u):\U\times\U\mapsto\R_{+}$ is any measurable cost function. $\bQ_Z$ is the marginal distribution of $Z$ when $X\sim P_X$ and $Z\sim \bQ(Z|X)$. (The proof is deferred to the Appendix. )
\end{theorem}

Therefore, optimizing over the objective on the r.h.s is equivalent to minimizing the discrepancy between the true label distribution ($P_Y$) and the output distribution $P_\bC$, thus the important classification features are embedded in the low-dimensional space.
This is the core idea of the paper, summarizing the high-dimensional data in a space of much lower dimension without losing important features for classification.
To implement the r.h.s objective, the constraint on $\bQ_Z$ can be relaxed by adding a penalty term. The final objective of OT-Classifier is:
\begin{align}
   \inf_{\bQ(Z|X)\in\mathcal{Q}}\E_{P_X}\E_{\bQ(Z|X)}\left\{\ell(f(X),\bC(Z))\right\}+\lambda\mathcal{D}(\bQ_Z,P_Z), 
\end{align}
where $\mathcal{Q}$ is any nonparametric set of probabilistic encoders, $\lambda>0$ is a hyper-parameter and $\mathcal{D}$ is an arbitrary divergence between $\bQ_Z$ and $P_Z$.

To estimate the divergences between $\bQ_Z$ and $P_Z$, we apply a GAN-based framework, fitting a discriminator to minimize the 1-Wasserstein distance between $\bQ_Z$ and $P_Z$: 
\begin{align*}
    W(\bQ_Z,P_Z)=\inf_{\Gamma\in\mP(\tilde{Z}\sim\bQ_Z,Z\sim P_Z)}\mathbb{E}_{(\tilde{Z},Z)\sim\Gamma}\|\tilde{Z}-Z\|.
\end{align*}
We have also tried the Jsensen-Shannon divergence, but as expected, Wasserstein distance provides more stable training and better results.
When training the framework, the weight clipping method proposed in Wasserstein GAN~\cite{arjovsky2017wasserstein} is applied to help stabilize the training of discriminator $\bD_\gamma$. 



\vspace{-5pt}
\section{Experiments}
\label{sec:exp}
In this section, we compare the performance of our proposed algorithm (OT-Classifier) with other state-of-the-art defense methods on several benchmark datasets:
\begin{itemize}[noitemsep,topsep=0pt,parsep=0pt,partopsep=0pt]
    \item MNIST~\cite{lecun1998mnist}: handwritten digit dataset, which consists of $60,000$ training images and $10,000$ testing images. Theses are $28\times 28$ black and white images in ten different classes.
    \item CIFAR10~\cite{krizhevsky2009learning}: natural image dataset, which contains $50,000$ training images and $10,000$ testing images in ten different classes. These are low resolution $32\times 32$ color images.
    \item STL10~\cite{coates2011analysis}: color image dataset similar to CIFAR10, but contains only $5,000$ training images and $8,000$ testing images in ten different classes. The images are of higher resolution $96\times 96$.
    \item Tiny Imagenet~\cite{deng2009imagenet}: a subset of Imagenet dataset. Tiny Imagenet has $200$ classes, and each class has $500$ training images, $50$ testing images, making it a challenging benchmark for defense task. The resolution of the images is $64\times 64$.
\end{itemize}

Various defense methods have been proposed to improve the robustness of deep neural networks. Here we compare our algorithm with state-of-the-art methods that are robust in white-box setting. Madry's adversarial training (\textbf{Madry's Adv}) is proposed in \cite{madry2017towards}, which has been recognized as one of the most successful defense method in white-box setting, as shown in~\cite{athalye2018obfuscated}. 

Random Self-Ensemble (\textbf{RSE}) method introduced by \cite{liu2017towards} adds stochastic components in the neural network, achieving similar performance to Madry's adversarial training algorithm. 

Another method we would like to compare with is \textbf{Defense-GAN}~\cite{samangouei2018defense}. It first trains a generative adversarial network to model the distribution of the training data. At inference time, it finds a close output to the input image and feed that output into the classifier. This process ``projects'' input images onto the range of GAN's generator, which helps remove the effect of adversarial perturbations. 
In \cite{samangouei2018defense}, the author demonstrated the performance of Defense-GAN on MNIST and Fashion-MNIST, so we will compare our method with Defense-GAN on MNIST.

Optimal transport classifier can be combined with other state-of-the-art defense methods. In general, Madry's adversarial training is more robust than RSE, so we combine OT-Classifier with adversarial training (\textbf{OT-CLA+Adv}) in our experiments.


\subsection{Evaluate Models Under White-box $l_\infty$-PGD Attack}
In this section, we evaluate the defense methods against $l_\infty$-PGD untargeted attack, which is one of the strongest white-box attack methods. 
Starting from $x^0=x_o$, PGD attack conducts
projected gradient descent iteratively to update the adversarial example: 
\begin{align*}
    x^{t+1}=\Pi_\epsilon\left\{x^t+\alpha\cdot\text{sign}\Big(\nabla_x\ell\big(\textbf{M}(x^t), y_o\big)\Big)\right\},
\end{align*}
where $\textbf{M}$ is the targeted model, $\Pi_{\epsilon}$ is the projection to the set $\{x|~\|x-x_o\|_\infty \le \epsilon\}$, $y_o$ is the label of $x_o$, and $\alpha$ is the step size. 
It is obvious that larger $\epsilon$ allows larger distortion of the original image.
Models are evaluated under different distortion level ($\epsilon$), and the larger the distortion the stronger the attack. Depending on the image scale and type, different datasets are sensitive to different strength of attack. 

\begin{figure*}
    \centering
    \includegraphics[width=0.245\textwidth]{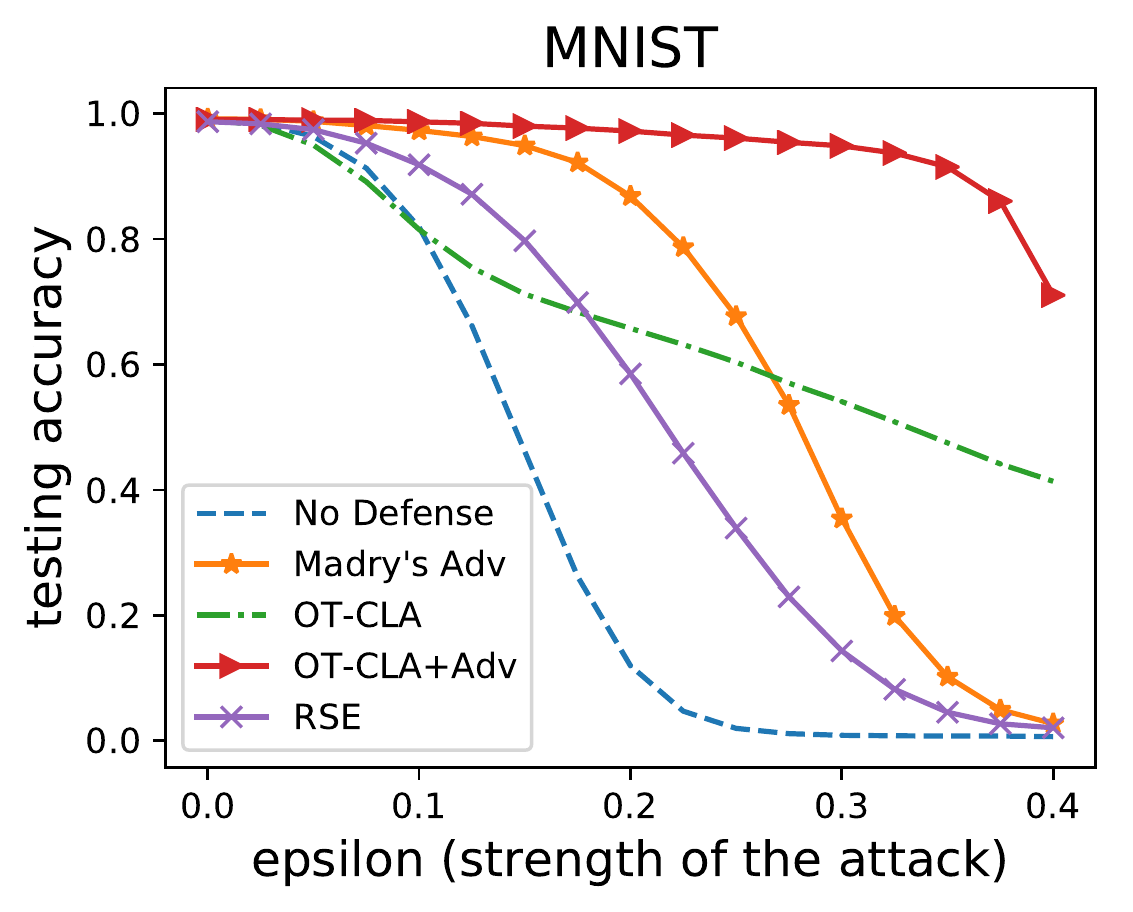}
    \includegraphics[width=0.245\textwidth]{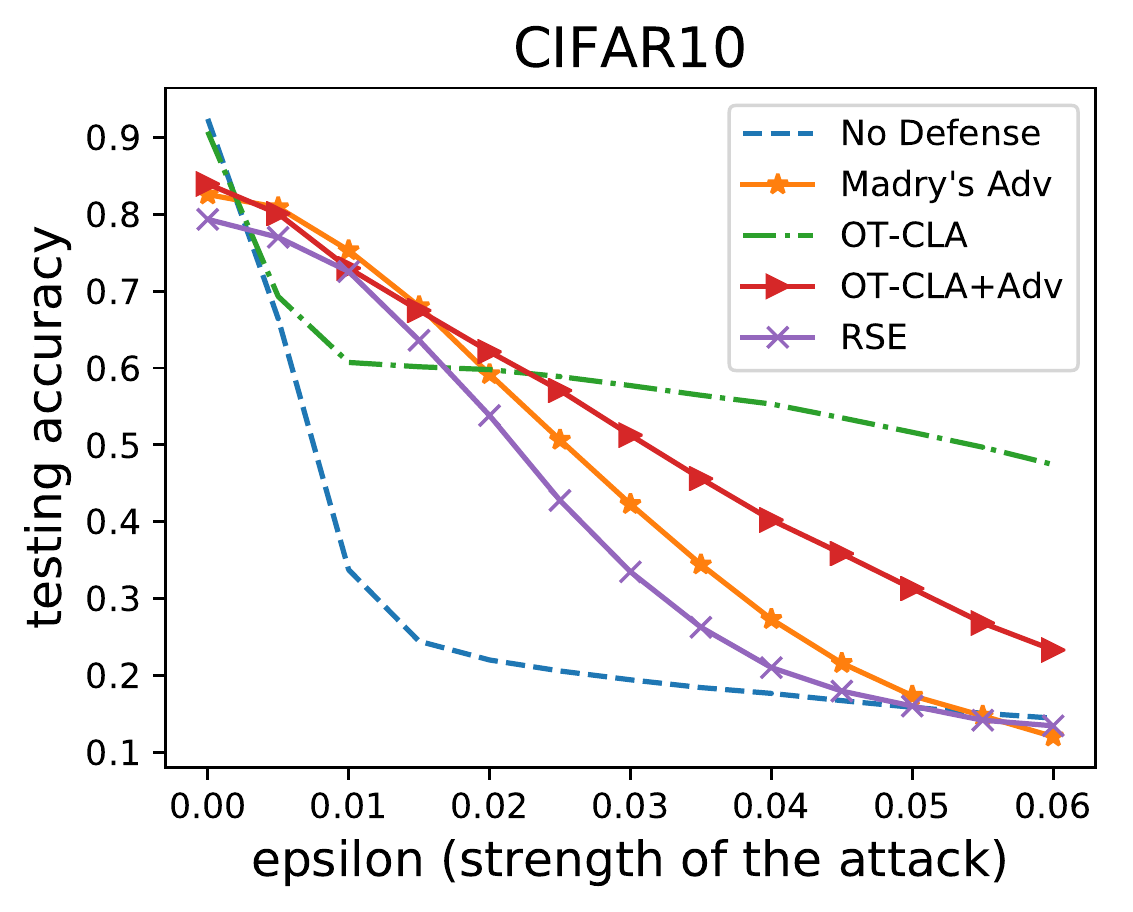}
    \includegraphics[width=0.245\textwidth]{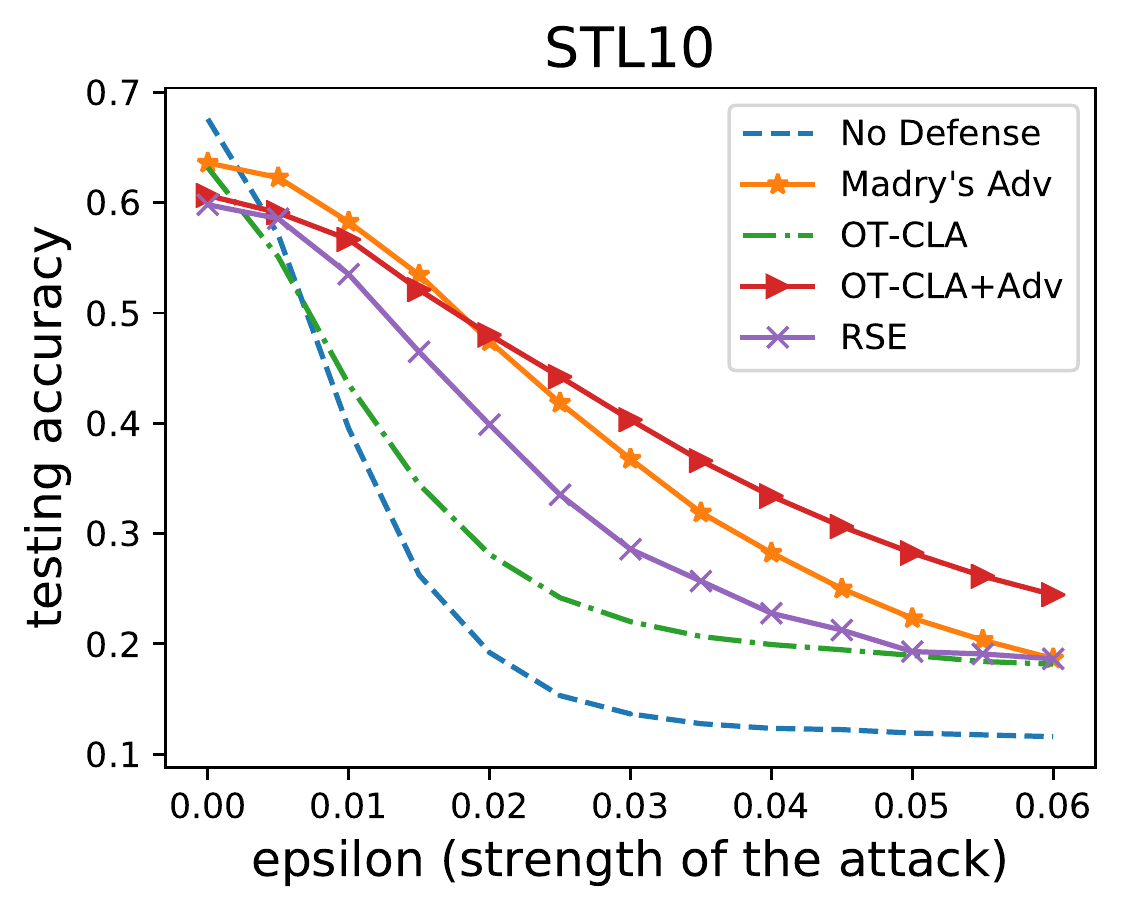}
    \includegraphics[width=0.245\textwidth]{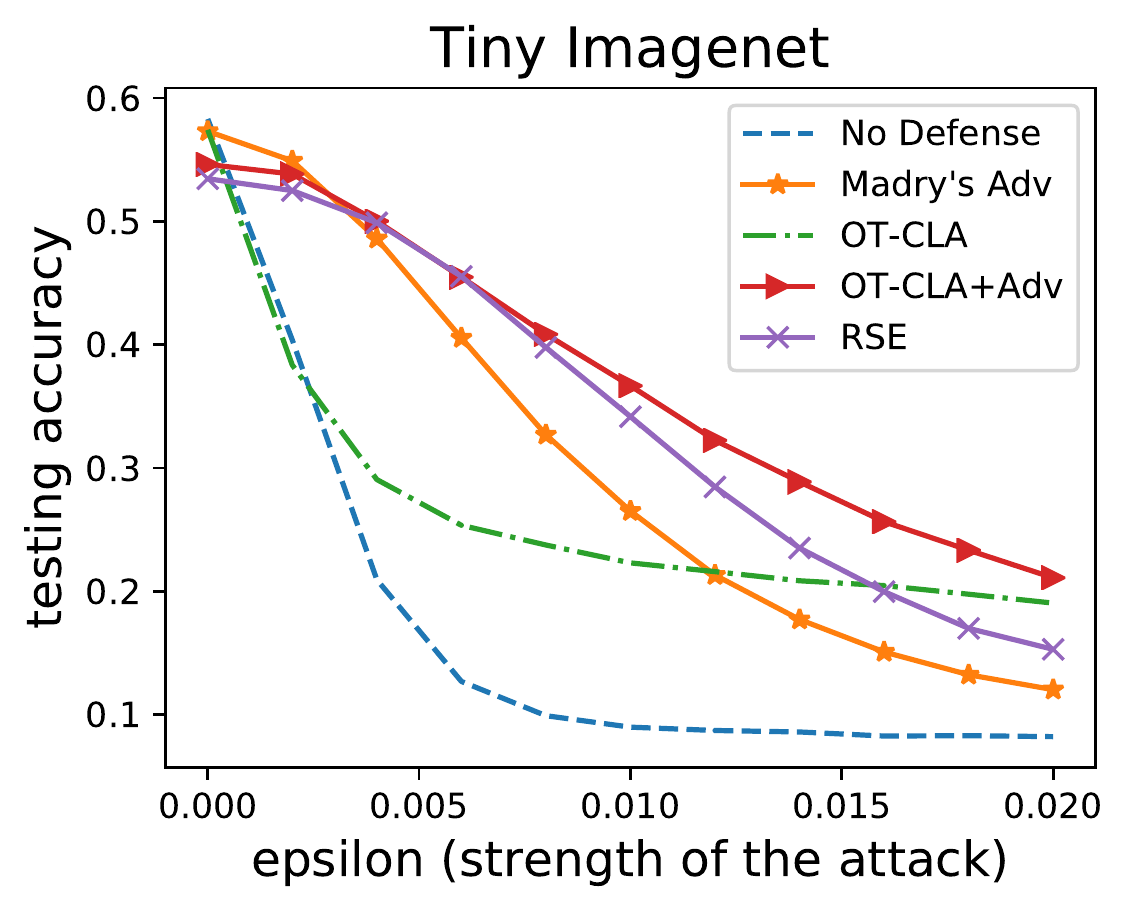}
    \caption{Testing accuracy under $l_\infty$-PGD attack on four different datasets: MNIST, CIFAR10, STL10 and Tiny Imagenet.}
    \label{fig:pgd_result}
\vspace{-10pt}
\end{figure*}

Models on MNIST are evaluated under distortion level from $0$ to $0.4$ by $0.025$. Models on CIFAR10 and STL10 are evaluated under $\epsilon \in [0,0.06,0.005]$. Models on Tiny Imagenet are evaluated under $\epsilon \in [0,0.02,0.002]$. As mentioned in the notation part, all the distortion levels are reported in the normalized $[0,1]$ space.
The experimental results are shown in Figure~\ref{fig:pgd_result}. To demonstrate the results more clearly, we show part of the results in Table~\ref{tab:acc_cmp}.

\begin{table}[htb]
    \centering
     \resizebox{8.3cm}{!}{
    \begin{tabular}{ccccccc}
    \toprule
    \textit{Data} & \textit{Defense} & \textit{0} & \textit{0.1} & \textit{0.2} & \textit{0.3} & \textit{0.4} \\\midrule
    \multirow{ 2}{*}{MNIST} & Adv. Training & \textbf{99.2} & 97.3 & 86.8 & 35.4 & 2.7 \\
     & OT-CLA+Adv & 99.1 & \textbf{98.7} & \textbf{97.2} & \textbf{94.9} & \textbf{71.1}\\\toprule\\\toprule
     
    \textit{Data} & \textit{Defense} & \textit{0} & \textit{0.015} & \textit{0.03} & \textit{0.045} & \textit{0.06}\\\midrule
    \multirow{ 2}{*}{CIFAR10} & Adv. Training & 82.6 & \textbf{68.0} & 42.3 & 21.6 & 12.0 \\
     & OT-CLA+Adv & \textbf{84.0} & 67.5 & \textbf{51.3} & \textbf{35.8} & \textbf{23.3} \\\midrule
     
     \multirow{2}{*}{STL10} & Adv. Training & \textbf{63.6} & 53.5 & 36.8 & 25.0 & 18.7 \\
     & OT-CLA+Adv & 60.7 & \textbf{52.1} & \textbf{40.3} & \textbf{30.6} & \textbf{24.5}\\\toprule\\\toprule
     
    \textit{Data} & \textit{Defense} & \textit{0} & \textit{0.004} & \textit{0.01} & \textit{0.016} & \textit{0.02}\\\midrule
    \multirow{2}{*}{Tiny Imagenet} & Adv. Training & \textbf{57.3} & 48.6 & 26.5 & 15.1 & 12.0 \\
    & OT-CLA+Adv &  54.6 &  \textbf{50.0} &  \textbf{36.7} & \textbf{25.6} & \textbf{21.1}\\
    \bottomrule
    \end{tabular}
    }
    \vspace{5pt}
    \caption{Testing accuracy ($\%$) under different strength of PGD attacks. The table shows the results of OT-CLA+Adv and Madry's adversarial training (Adv. Training). The better accuracy is marked in \textbf{bold}.}
    \label{tab:acc_cmp}
\end{table}

Based on Figure~\ref{fig:pgd_result} and Table~\ref{tab:acc_cmp}, we can see that OT-Classifier can improve the robustness of deep neural networks. Compare the performance of OT-Classifier with the performance of model without defense method, we can see that OT-Classifier is much more robust than the model with no defense method on all benchmark datasets. Besides, when the distortion level ($\epsilon$) is large, OT-Classifier tends to perform better than other state-of-the-art defense methods on MNIST, CIFAR10 and Tiny Imagenet. This phenomenon is obvious on CIFAR10 and it even performs better than OT-CLA+Adv when the attack strength is strong. 

In general, OT-Classifier combined with adversarial training (OT-CLA+Adv) is the most robust one on a variety of datasets. Though, on some datasets, when there is no attack, the testing accuracy of OT-CLA+Adv are slightly worse than Madry's adversarial training. 


We also compare Defense-GAN with our method OT-CLA+Adv on MNIST. Both methods are evaluated against the $l_2$-C$\&$W untargeted attack, one of the strongest white-box attack proposed in \cite{carlini2017towards}.  Defense-GAN is evaluated using the method proposed in \cite{athalye2018obfuscated}, and the code is available on github \footnote{Publicly available at \url{https://github.com/anishathalye/obfuscated-gradients/tree/master/defensegan}}. OT-CLA+Adv is evaluated against $l_2$-C$\&$W untargeted attack with the same hyper-parameter values as those used in the evaluation of Defense-GAN. The results under $l_2\leq0.005$ threshold are shown in Table~\ref{tab:defense_gan}.

\begin{table}[H]
\vspace{-5pt}
    \centering
    \begin{tabular}{|c|c|}\hline
      Method   & Testing Accuracy \\ \hline
      Defense-GAN   & 55.0 \\
      OT-CLA+Adv & 99.1 \\ \hline
    \end{tabular}
    \vspace{5pt}
    \caption{Testing accuracy ($\%$) of two defense methods under C$\&$W attack with $l_2\leq0.005$.
    }
    \label{tab:defense_gan}
\vspace{-10pt}
\end{table}

Based on Table~\ref{tab:defense_gan}, OT-CLA+Adv is much more robust than Defense-GAN under the $l_2\leq0.005$ threshold. 

\subsection{Evaluate the Effect of Discriminator}
OT-Classifier framework consists of three parts, and the classification task is done by the encoder $\bQ_\phi$ and classifier $\bC_\tau$. Without the discriminator part, the encoder can also project the input images to a low-dimensional space. However, arbitrarily projecting the images to a low-dimensional space with only the encoder part can not improve the robustness of the model. In contrast, sometimes it even decreases the robustness of the model. 

\begin{figure}
    \centering
    \includegraphics[width=0.235\textwidth]{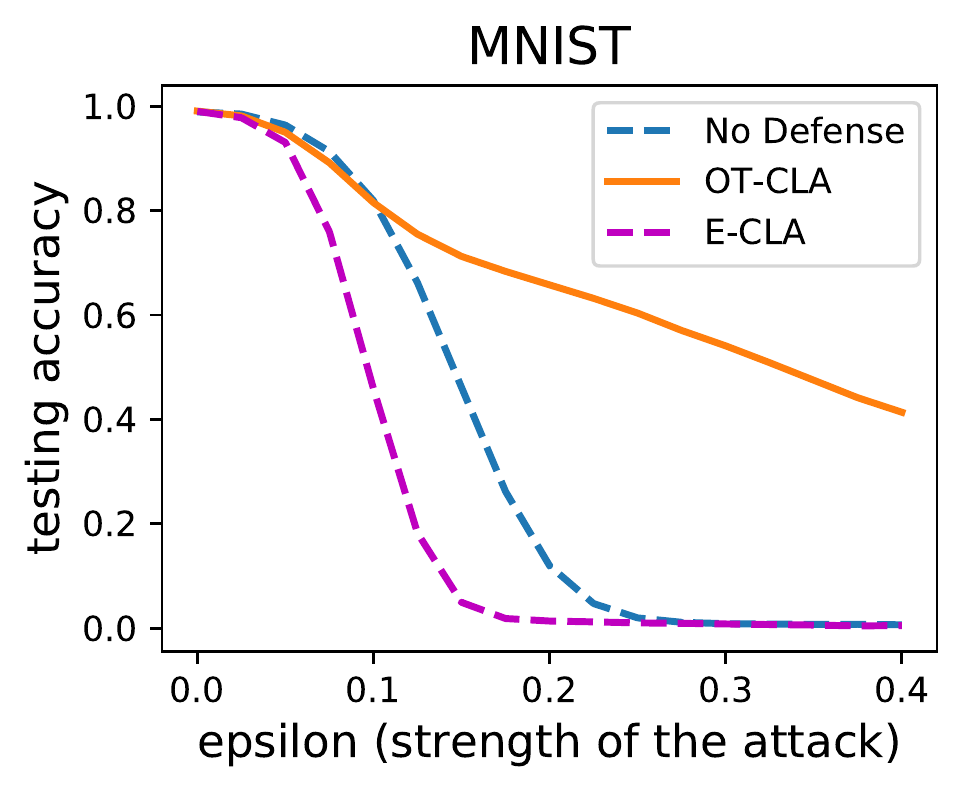}
    \includegraphics[width=0.235\textwidth]{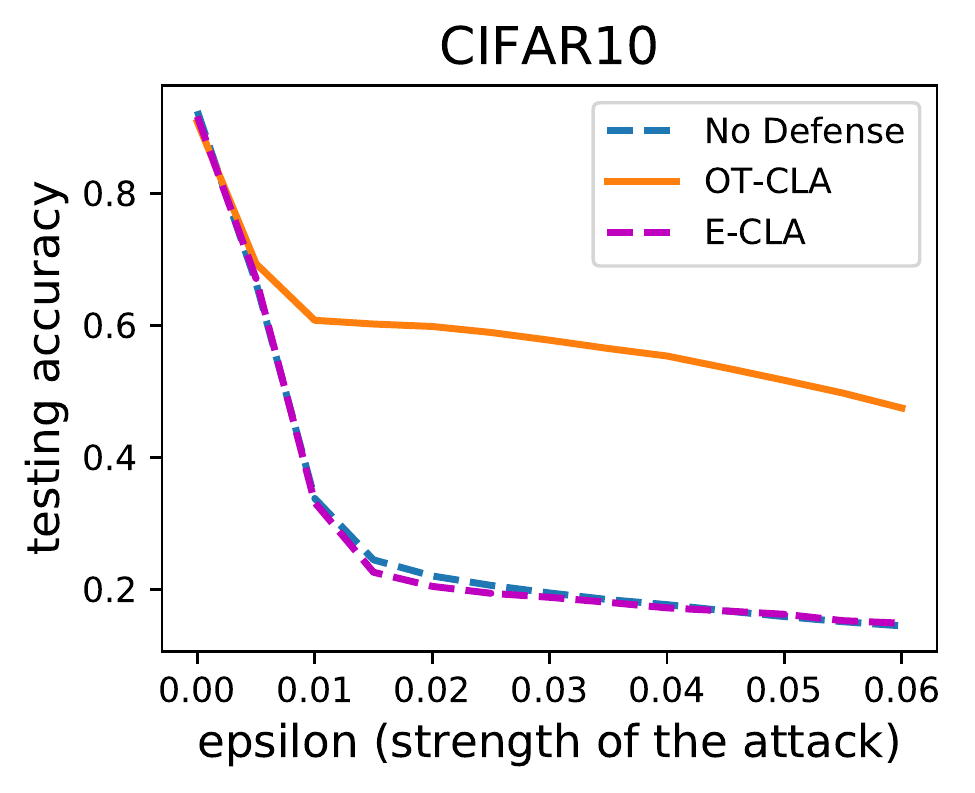}
    \includegraphics[width=0.235\textwidth]{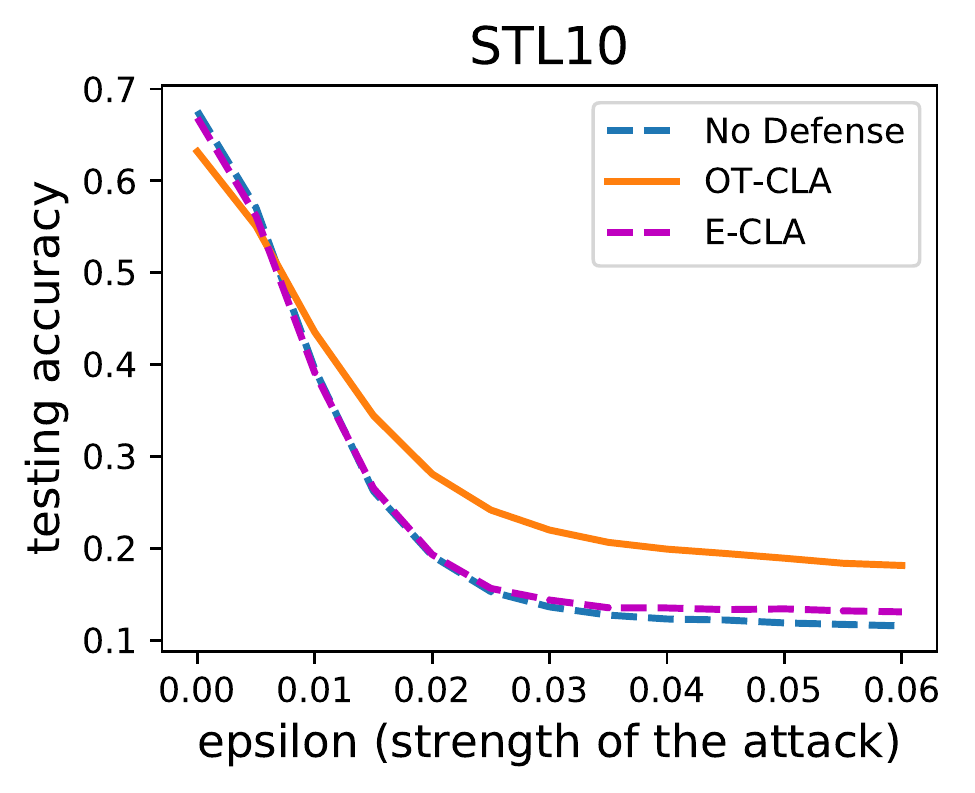}
    \includegraphics[width=0.235\textwidth]{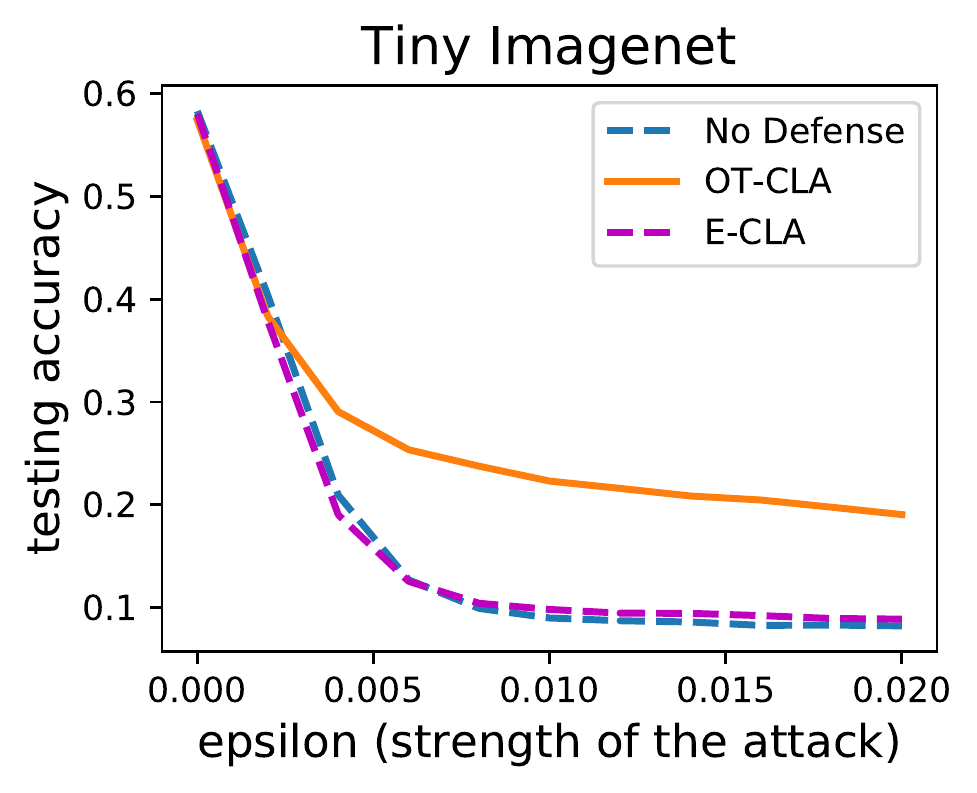}
    \caption{Testing accuracy of E-CLA and OT-Classifier under $l_\infty$-PGD attack on four different datasets: MNIST, CIFAR10, STL10 and Tiny Imagenet. We adopt the same encoder and classifier structures for the two models.}
    \label{fig:enccla}
\vspace{-10pt}
\end{figure}

To show that arbitrarily projecting the input images to a low-dimensional space can not improve the robustness, we fit a framework with only the encoder and classifier part (\textbf{E-CLA}), where the encoder and classifier have the same structures as in OT-Classifier, and compare E-CLA with the OT-Classifier framework. The results are shown in Figure~\ref{fig:enccla}. 

Based on Figure~\ref{fig:enccla}, we can observe that OT-Classifier is much more robust than just the encoder and classifier structure on MNIST, CIFAR10 and Tiny Imagenet. It is also more robust on STL10 but not that much. The reason might be that there are only $5,000$ training images in STL10 and the resolution is $96\times 96$. Therefore, it is harder to learn a good embedding with limited amount of images.  However, even when the number of training images is limited, OT-Classifier is still much more robust than the E-CLA structure. This observation demonstrates that OT-Classifier is able to learn a robust embedding. Notice that the performance of E-CLA structure is similar to the performance of model without defense method on CIFAR10, STL10 and Tiny Imagenet, and worse on MNIST, which means the robustness of OT-Classifier does not come from the structure design. 


\subsection{Dimension of Embedding Space}

\begin{figure*}
    \centering
    \includegraphics[width=0.25\textwidth]{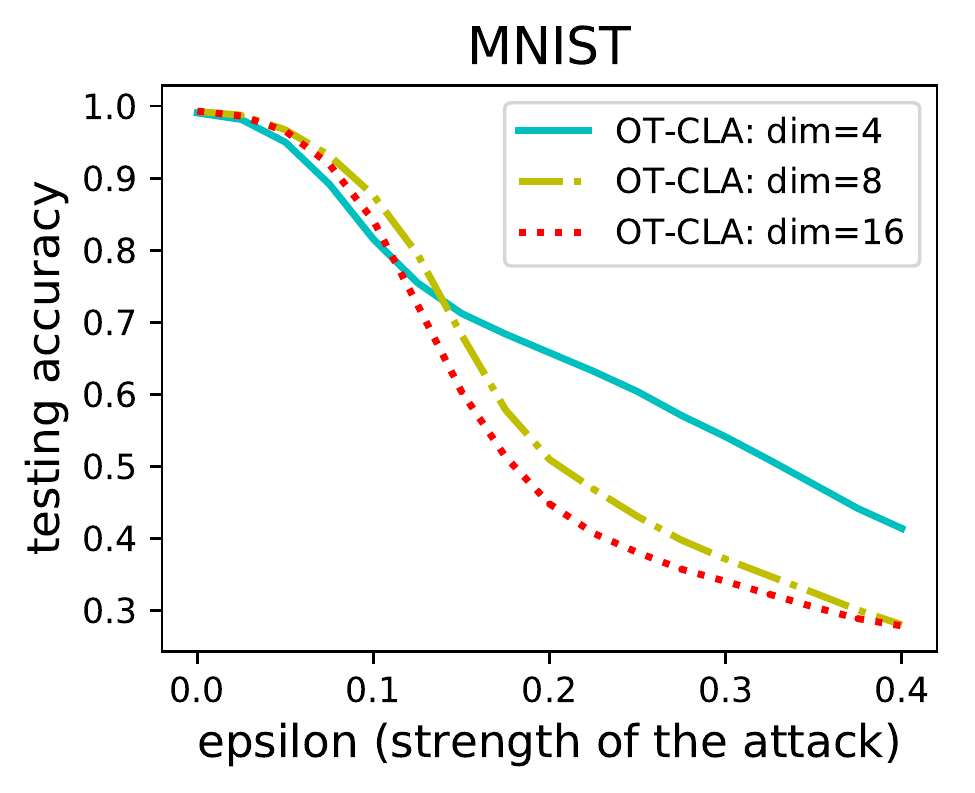}
    \includegraphics[width=0.25\textwidth]{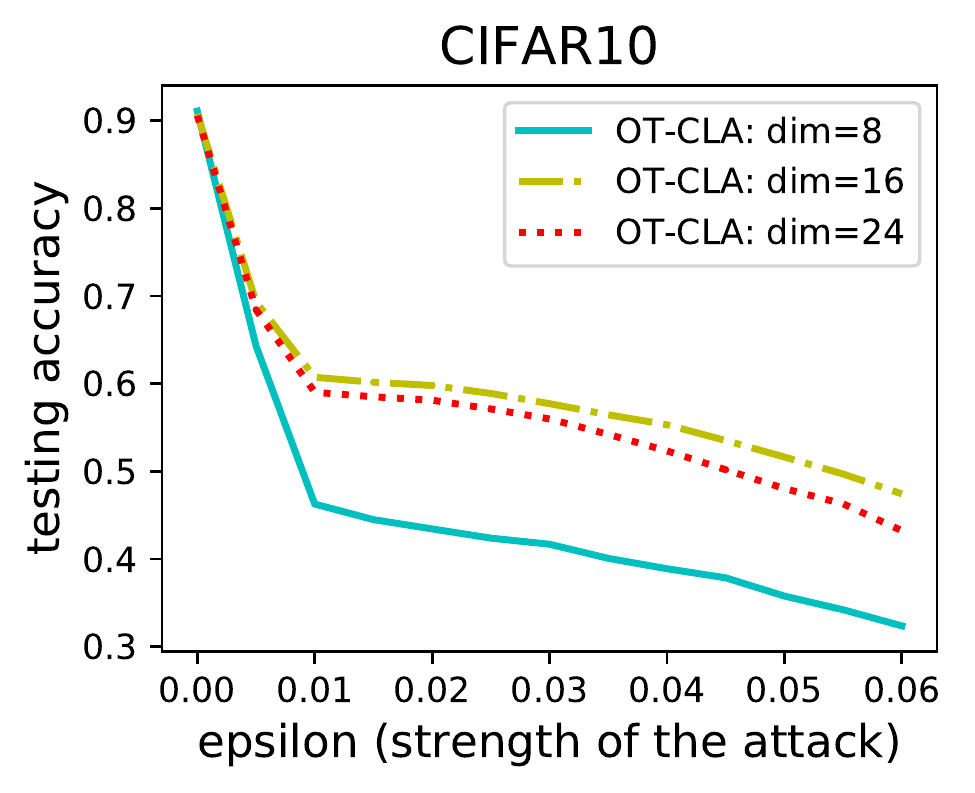}
    \includegraphics[width=0.25\textwidth]{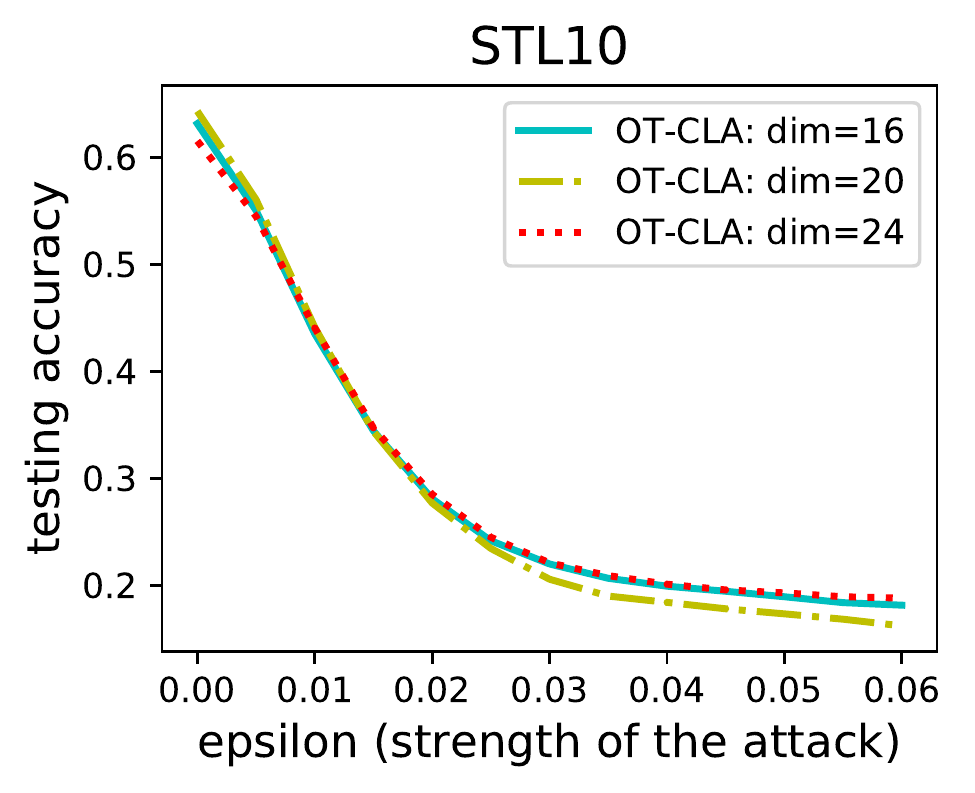}
    \caption{Testing accuracy of models with different embedding dimensions under $l_\infty$-PGD attack.}
    \label{fig:dim}
\end{figure*}

One important hyper-parameter for the OT-Classifier is the dimension of the embedding space. If the dimension is too small, important features are ``collapsed'' onto the same dimension, and if the dimension is too large, the projection will not extract useful information, which results in too much noise and instability. The maximum likelihood estimation of intrinsic dimension proposed in \cite{levina2005maximum}\footnote{Code publicly available at \url{https://github.com/OFAI/hub-toolbox-python3}} is used to calculate the intrinsic dimension of each image dataset, serving as a guide for selecting the embedding dimension. The sample size used in calculating the intrinsic dimension is $1,000$, and changing the sample size does not influence the results much. Based on the intrinsic dimension calculated by \cite{levina2005maximum}, we test several different values around the suggested intrinsic dimension and evaluate the models against $l_\infty$-PGD attack. The experimental results are shown in Figure~\ref{fig:dim}.

The final embedding dimension is chosen based on robustness, number of parameters, and testing accuracy when there is no attack. The final embedding dimensions and suggested intrinsic dimensions are shown in Table~\ref{tab:in_dim}. 

\begin{table}[H]
    \centering
     \resizebox{8.3cm}{!}{
    \begin{tabular}{|c|c|c|c|}
    \hline
    Data &  Data dim. & Intrinsic dim. & Embedding dim.\\ \hline
    MNIST &  $1\times28\times 28$ & 13 & 4\\
    CIFAR10 &  $3\times32\times32$ & 17 & 16\\
    STL10 &  $3\times96\times96$ & 20 & 16\\
    Tiny Imagenet &  $3\times64\times64$ & 19 & 20\\ \hline
    \end{tabular}
    }
    \vspace{5pt}
    \caption{Pixel space dimension, intrinsic dimension calculated by~\cite{levina2005maximum}, and final embedding dimension used.}
    \label{tab:in_dim}
\end{table}

Based on Figure~\ref{fig:dim}, the embedding dimension close to the calculated intrinsic dimension usually offers better results except on MNIST. One explanation may be that MNIST is a simple handwritten digit dataset, so performing classification on MNIST may not require that many dimensions.

\subsection{Embedding Visualization}
\begin{figure*}
    \centering
    \includegraphics[width=0.245\textwidth]{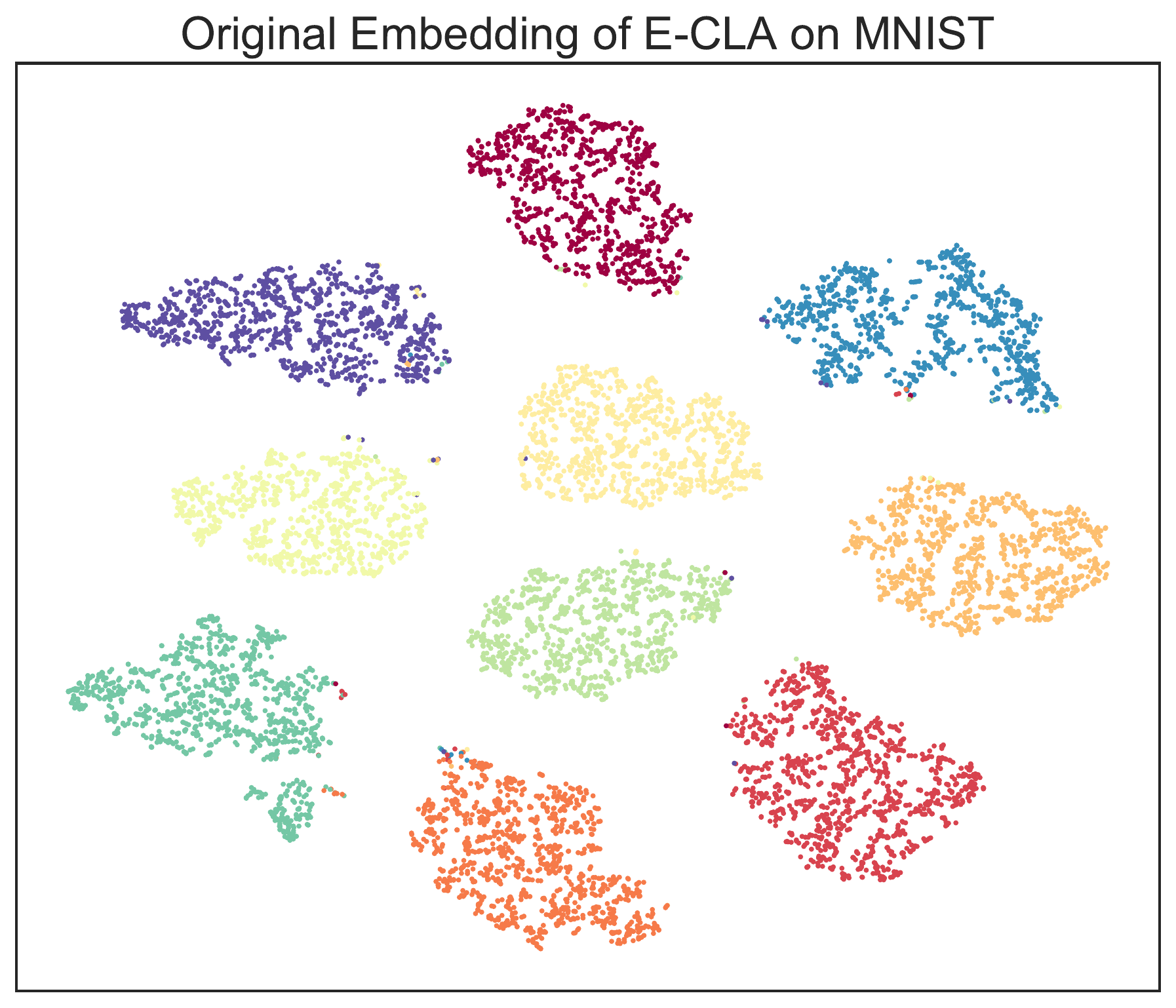}
    \includegraphics[width=0.245\textwidth]{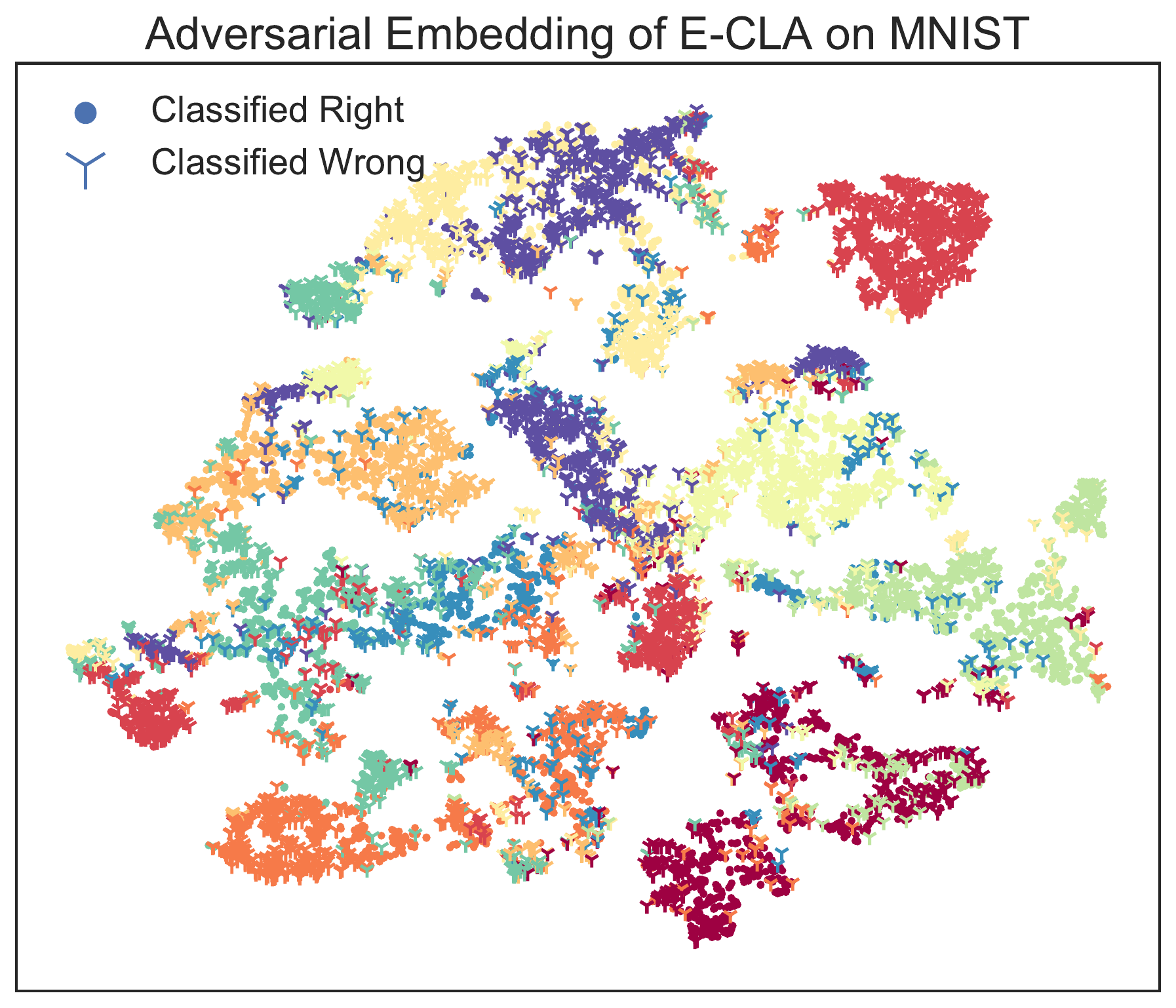}
    \includegraphics[width=0.245\textwidth]{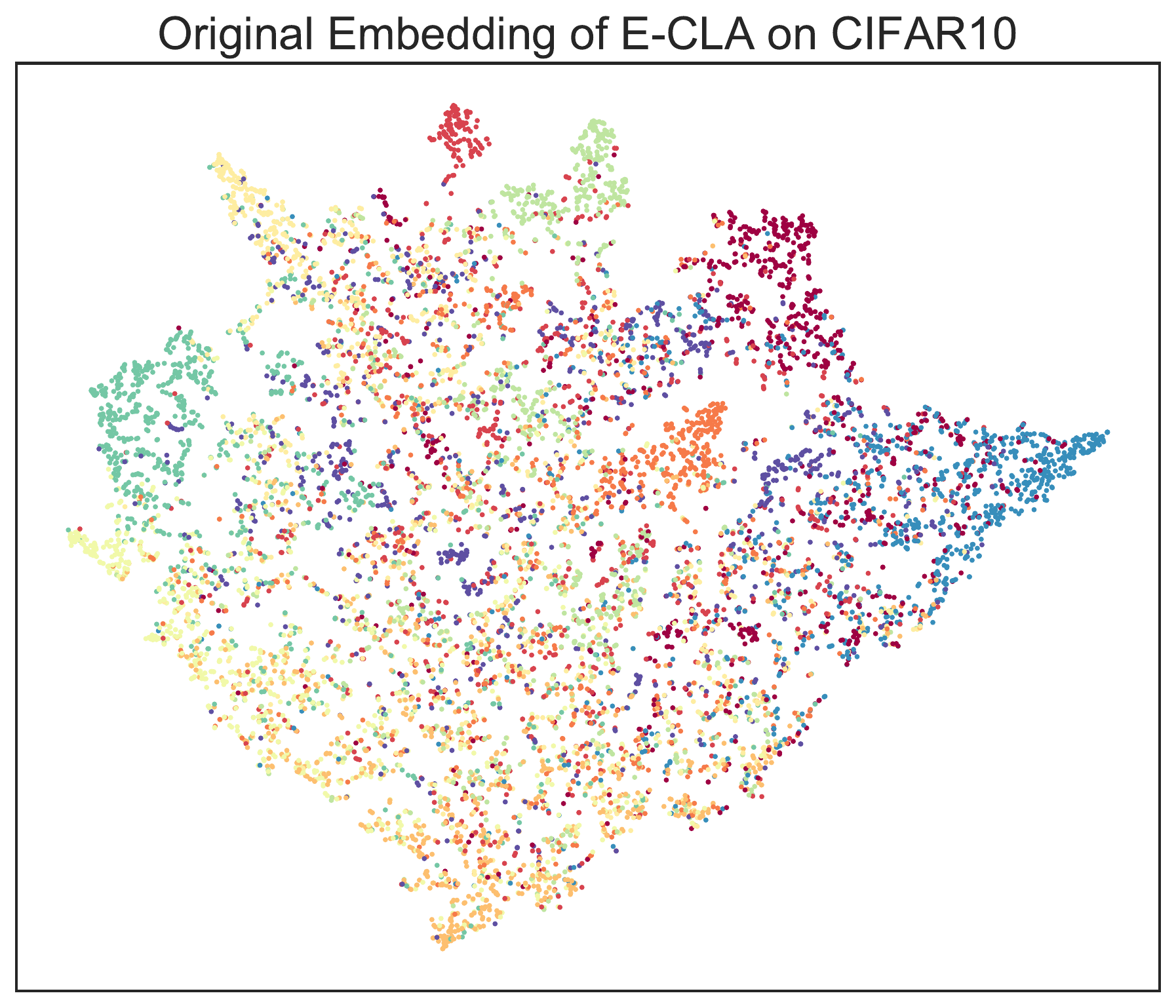}
    \includegraphics[width=0.245\textwidth]{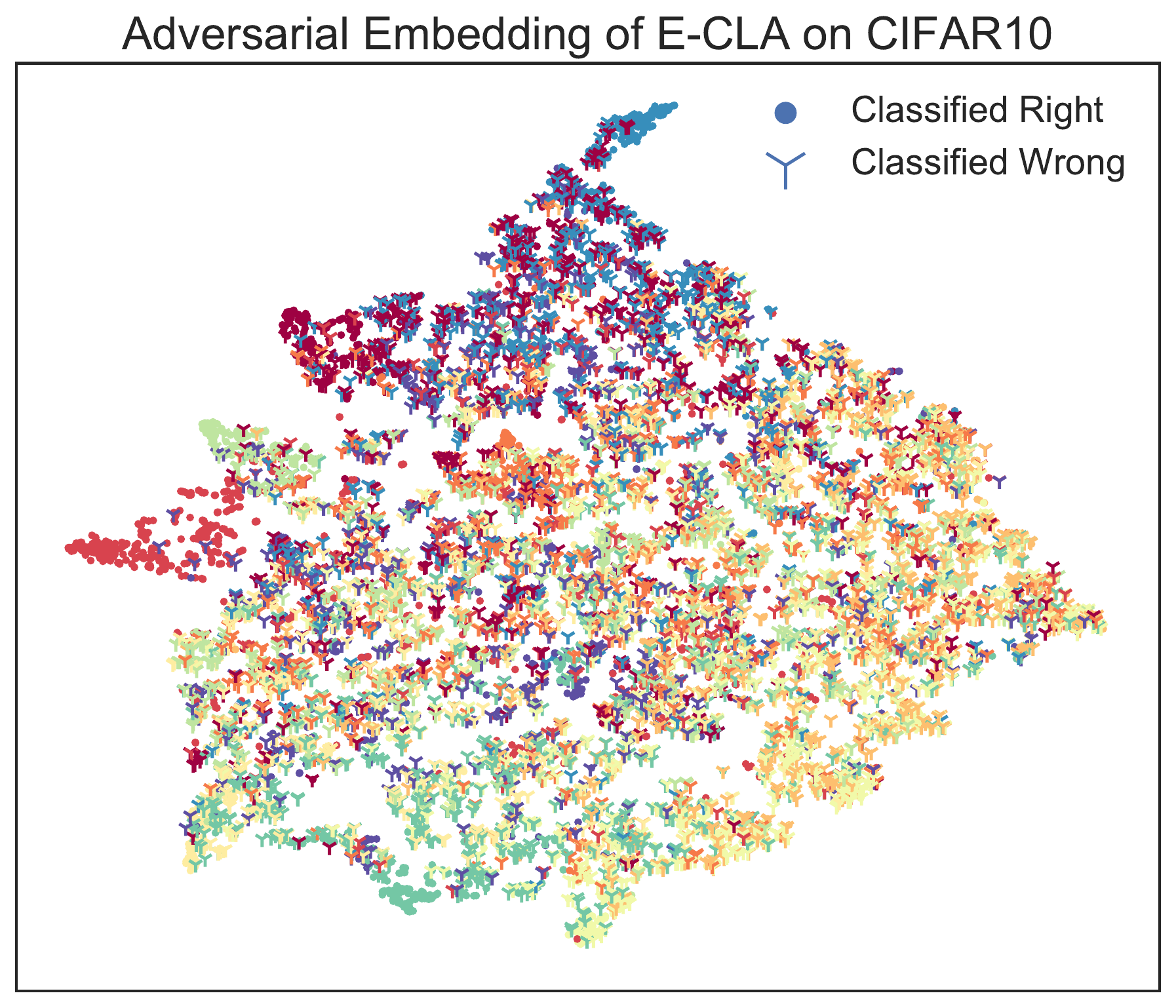}
    
    \includegraphics[width=0.245\textwidth]{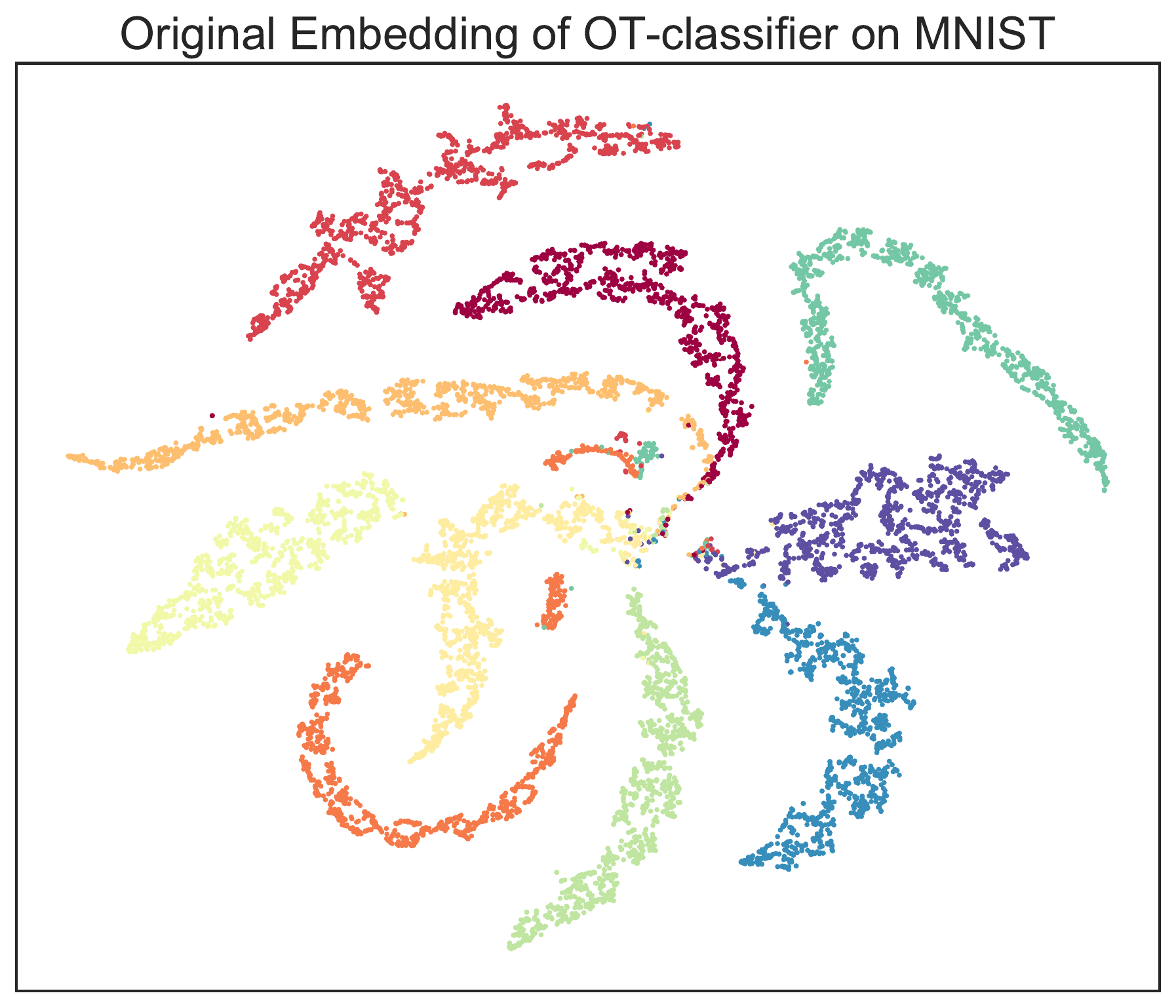}
    \includegraphics[width=0.245\textwidth]{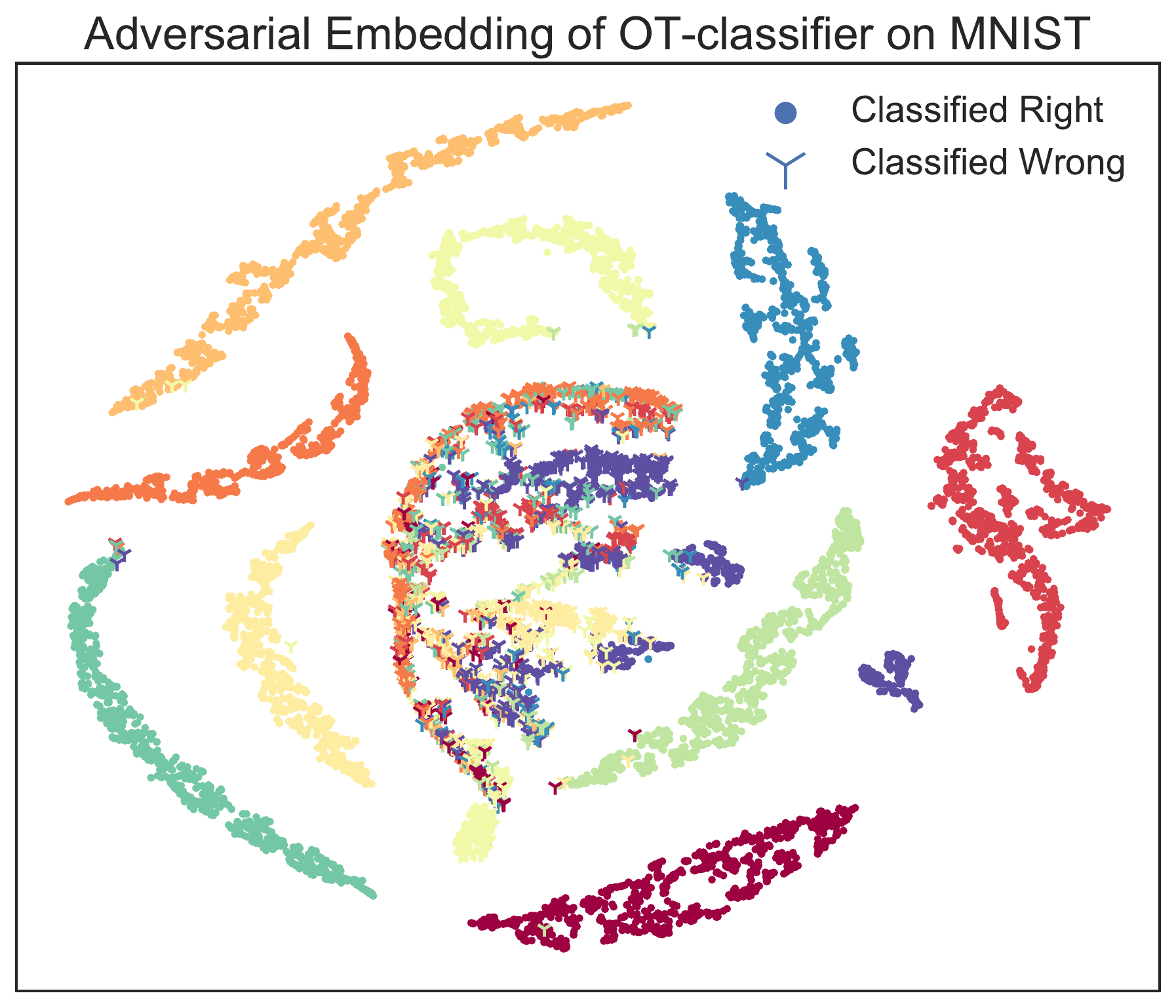}
    \includegraphics[width=0.245\textwidth]{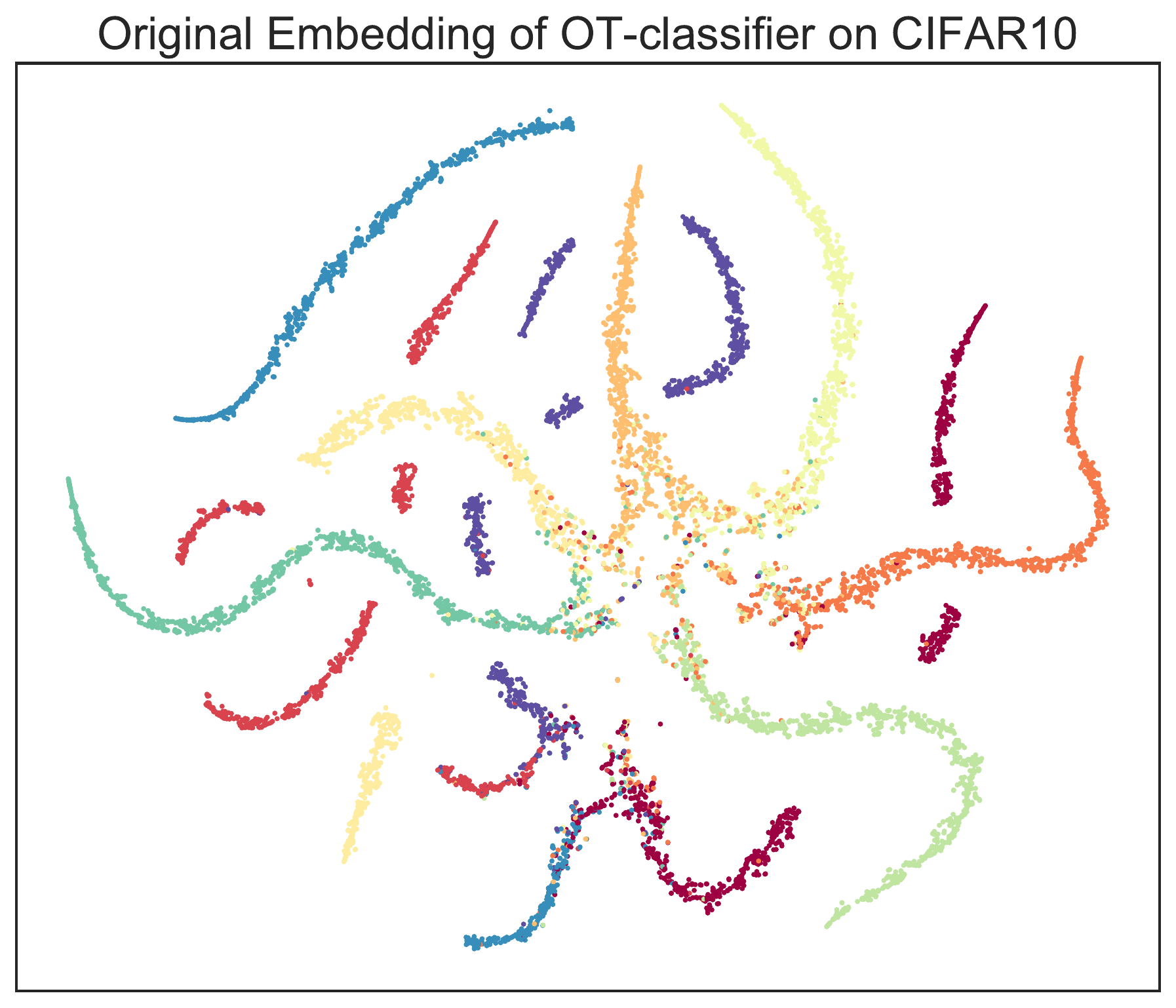}
    \includegraphics[width=0.245\textwidth]{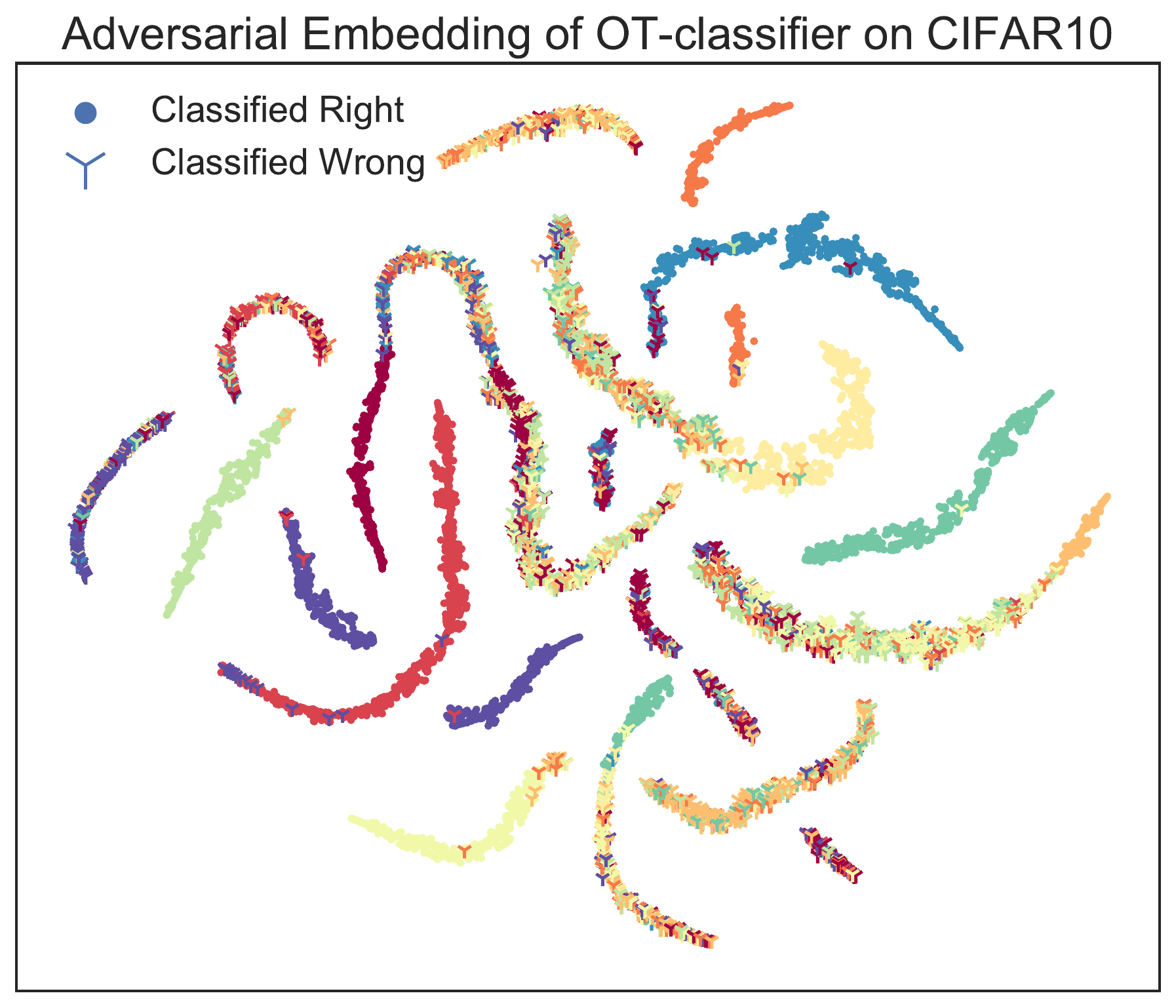}
    \caption{2D embeddings for E-CLA and OT-Classifier on MNIST and CIFAR10. See larger plots in Supplementary.}
    \label{fig:embed}
    \vspace{-10pt}
\end{figure*}

In this section, we compare the embedding learned by Encoder+Classifier structure (\textbf{E-CLA}) and the embedding learned by OT-Classifier on several datasets. We first generate embedding of testing data using the encoder ($\tilde{z} = \bQ_\phi(x)$), then project the embedding points ($\tilde{z}$) to 2-D space by tSNE\cite{maaten2008visualizing}. Then we generate adversarial images ($x_{adv}$) against E-CLA and OT-Classifier using $l_\infty$-PGD attack. The adversarial embedding is generated by feeding the adversarial images into the encoder ($\tilde{z}_{adv} = \bQ_\phi(x_{adv})$). Finally, we project the adversarial embedding points ($\tilde{z}_{adv}$) to 2-D space. 
The results are shown in Figure~\ref{fig:embed}. The plots in the first row are embedding visualization plots for E-CLA, and the plots in the second row are the embedding visualization plots for OT-Classifier. In adversarial embedding visualization plots, the misclassified point is marked as ``down triangle'', which means the PGD attack successfully changed the prediction, and the correctly classified point is marked as ``point'', which means the attack fails. 


Based on Figure~\ref{fig:embed}, we can see that E-CLA can learn a good embedding on legitimate images of MNIST. Embedding points for different classes are separated on the 2D space, but under adversarial attack, some embedding points of different classes are mixed together. However, OT-Classifier can generate good separated embeddings on both legitimate and adversarial images. On CIFAR10, the E-CLA can not generate good separated embeddings on either legitimate images or adversarial images, while OT-Classifier can generate good separated embeddings for both.



\vspace{-5pt}
\section{Conclusion}
In this paper, we propose a new defense framework, OT-Classifier, which projects the input images to a low-dimensional space to remove adversarial perturbation and stabilize the model through minimizing the discrepancy between the true label distribution and the framework output distribution. We empirically show that OT-CLA+Adv is much more robust than other state-of-the-art defense methods on several benchmark datasets. Future work will include further exploration of the low-dimensional space to improve the robustness of deep neural network.
\vspace{-5pt}
\section{Appendix}
\subsection{Proof of Theorem 1}
The proof of Theorem 1 is adapted from the proof of Theorem 1 in \cite{tolstikhin2017wasserstein}. Consider certain sets of joint probability distributions of three random variables $(X,U,Z)\in\X\times\U\times\Z$. $X$ can be taken as the input images, $U$ as the output of the framework, and $Z$ as the latent codes. $P_{\bC,Z}(U,Z)$ represents a joint distribution of a variable pair $(U,Z)$, where $Z$ is first sampled from $P_Z$ and then $U$ from $P_\bC(U|Z)$. $P_\bC$ defined in~(\ref{eq:p_c}) is the marginal distribution of $U$ when $(U,Z)\sim P_{\bC,Z}$. 

The joint distributions $\Gamma(X,U)$ or couplings between values of $X$ and $U$ can be written as $\Gamma(X,U)=\Gamma(U|X)P_X(X)$ due to the marginal constraint. $\Gamma(U|X)$ can be decomposed into an encoding distribution $Q(Z|X)$ and the generating distribution $P_\bC(U|Z)$, and Theorem 1 mainly shows how to factor it through $Z$.

In the first part, we will show that if $P_\bC(U|Z)$ are Dirac measures, we have
\begin{align*}
    &\inf_{\Gamma\in\mP(X\sim P_X, U\sim P_\bC)}\E_{(X,U)\sim\Gamma}\left\{\ell(f(X),U)\right\}\\
    &=\inf_{\Gamma\in\mP_{X,U}}\E_{(X,U)\sim\Gamma}\left\{\ell(f(X),U)\right\}, \numberthis \label{eq:pf1}
\end{align*}
where $\mP(X\sim P_X, U\sim P_\bC)$ denotes the set of all joint distributions of $(X,U)$ with marginals $P_X, P_\bC$, and likewise for $\mP(X\sim P_X, Z\sim P_Z)$. The set of all joint distributions of $(X,U,Z)$ such that $X\sim P_X$, $(U,Z)\sim P_{\bC,Z}$, and $(U\independent X)|Z$ are denoted by $\mP_{X,U,Z}$. $\mP_{X,U}$ and $\mP_{X,Z}$ denote the sets of marginals on $(X,U)$ and $(X,Z)$ induced by $\mP_{X,U,Z}$. 

From the definition, it is clear that $\mP_{X,U}\subseteq\mP(P_X,P_\bC)$. Therefore, we have
\begin{align*}
    &\inf_{\Gamma\in\mP(X\sim P_X, U\sim P_\bC)}\E_{(X,U)\sim\Gamma}\left\{\ell(f(X),U)\right\}\\
    &\leq \inf_{\Gamma\in\mP_{X,U}}\E_{(X,U)\sim\Gamma}\left\{\ell(f(X),U)\right\}, \numberthis \label{eq:pf2}    
\end{align*}
The identity is satisfied if $P_\bC(U|Z)$ are Dirac measures, such as $U=\bC(Z)$. This is proved by the following Lemma in \cite{tolstikhin2017wasserstein}.
\vspace{-5pt}
\begin{lemma}
$\mP_{X,U}\subseteq\mP(P_X,P_\bC)$ with identity if $P_\bC(U|Z=z)$ are Dirac for all $z\in\Z$. (see details in~\cite{tolstikhin2017wasserstein}.)
\end{lemma}

In the following part, we show that 
\begin{align*}
&\inf_{\Gamma\in\mP_{X,U}}\E_{(X,U)\sim\Gamma}\left\{\ell(f(X),U)\right\}\\
&=\inf_{\bQ:\bQ_Z=P_Z}\E_{P_X}\E_{\bQ(Z|X)}\left\{\ell(f(X),\bC(Z))\right\}. \numberthis \label{eq:pf3}    
\end{align*}
Based on the definition, $\mP(P_X,P_\bC)$, $\mP_{X,U,Z}$ and $\mP_{X,U}$ depend on the choice of conditional distributions $P_\bC(U|Z)$, but $\mP_{X,Z}$ does not. It is also easy to check that $\mP_{X,Z}=\mP(X\sim P_X,Z\sim P_Z)$. The tower rule of expectation, and the conditional independence property of $\mP_{X,U,Z}$ implies
\begin{align*}
    &\inf_{\Gamma\in\mP_{X,U}}\E_{(X,U)\sim\Gamma}\left\{\ell(f(X),U)\right\}\\
    &=\inf_{\Gamma\in\mP_{X,U,Z}}\E_{(X,U,Z)\sim\Gamma}\left\{\ell(f(X),U)\right\}\\
    &=\inf_{\Gamma\in\mP_{X,U,Z}}\E_{P_Z}\E_{X\sim P(X|Z)}\E_{U\sim P(U|Z)}\left\{\ell(f(X),U)\right\}\\
    &=\inf_{\Gamma\in\mP_{X,U,Z}}\E_{P_Z}\E_{X\sim P(X|Z)}\left\{\ell(f(X),\bC(Z))\right\}\\
    &=\inf_{\Gamma\in\mP_{X,Z}}\E_{(X,Z)\sim\Gamma}\left\{\ell(f(X),\bC(Z))\right\}\\
    &=\inf_{\bQ:\bQ_Z=P_Z}\E_{P_X}\E_{\bQ(Z|X)}\left\{\ell(f(X),\bC(Z))\right\} \numberthis \label{eq:pf4} 
\end{align*}

Finally, since $Y=f(X)$, it is easy to get
\begin{align*}
    &\inf_{\Gamma\in\mP(Y\sim P_Y, U\sim P_\bC)}\E_{(Y,U)\sim\Gamma}\left\{\ell(Y,U)\right\}\\
    &=\inf_{\Gamma\in\mP(X\sim P_X, U\sim P_\bC)}\E_{(X,U)\sim\Gamma}\left\{\ell(f(X),U)\right\} \numberthis \label{eq:pf5} 
\end{align*}
Now (\ref{eq:pf1}), (\ref{eq:pf3}) and (\ref{eq:pf5}) are proved and the three together prove Theorem 1. 

{\small
\bibliographystyle{ieee}
\bibliography{egbib}
}

\clearpage
    \begingroup
    \let\clearpage\relax 
    \onecolumn 
\section*{Supplementary}
\FloatBarrier
\vspace{15pt}
\subsection*{A. Tuning Epsilon}
\vspace{15pt}
Epsilon ($\epsilon$) is an important hyper-parameter for adversarial training. When doing Madry's adversarial training, we test the model robustness with different $\epsilon$ and choose the best one. The experiment results are shown in Figure~\ref{fig:epsilon}.
\begin{figure}
\vspace{-2cm}
    \centering
    \includegraphics[width=0.32\textwidth]{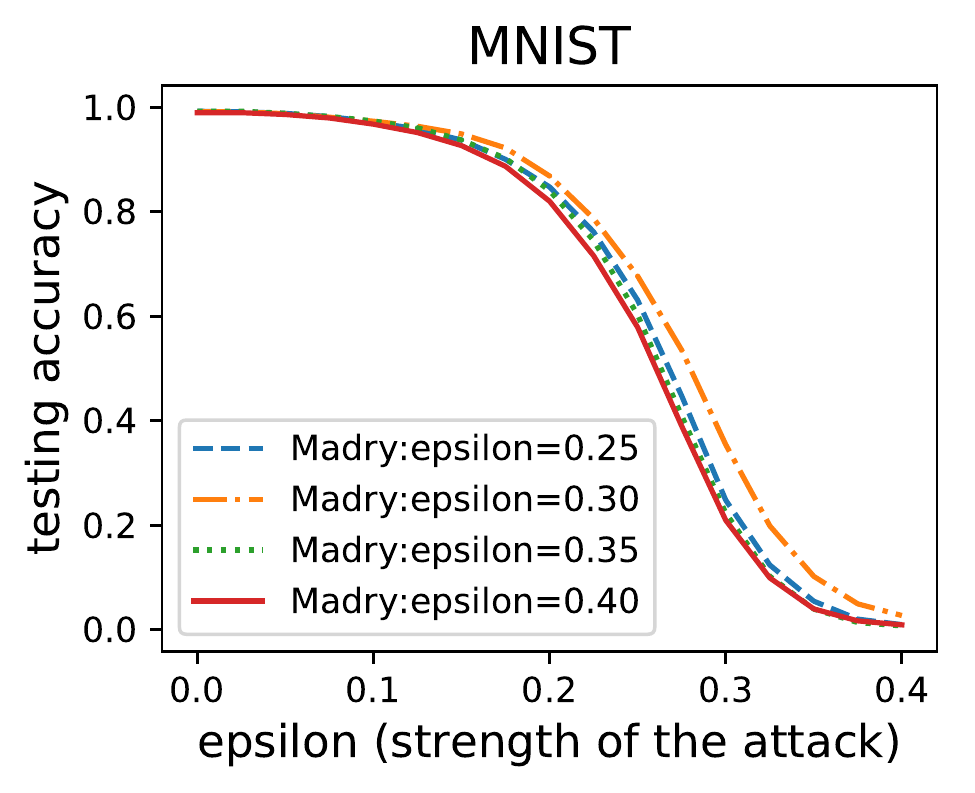}
    \includegraphics[width=0.32\textwidth]{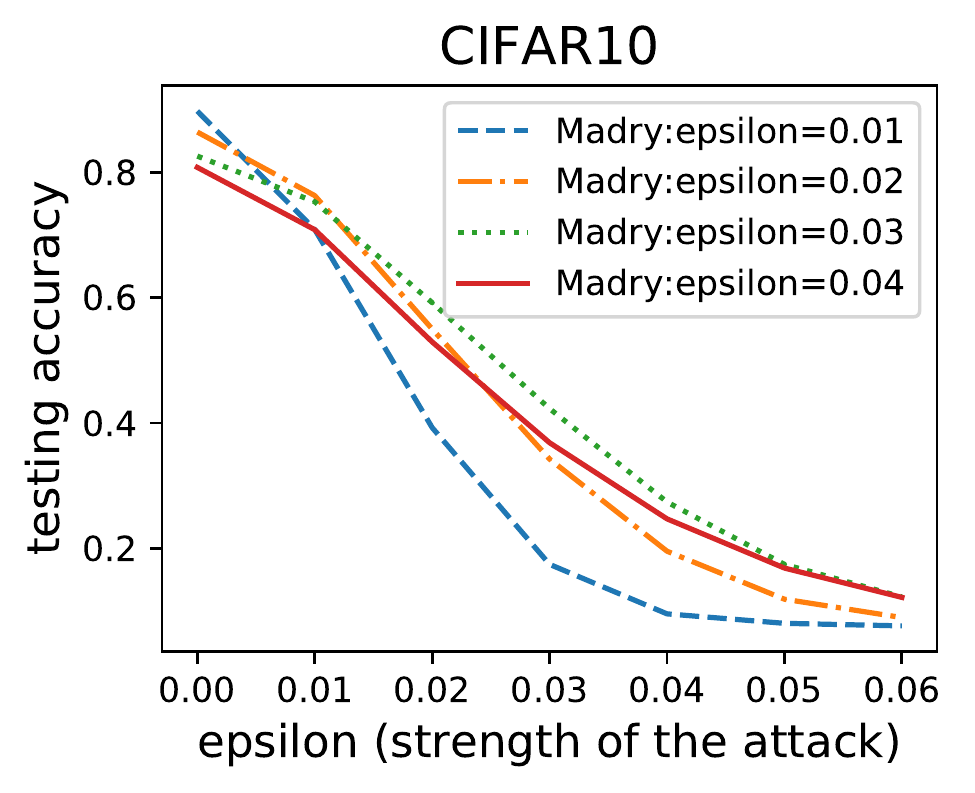}
    \includegraphics[width=0.32\textwidth]{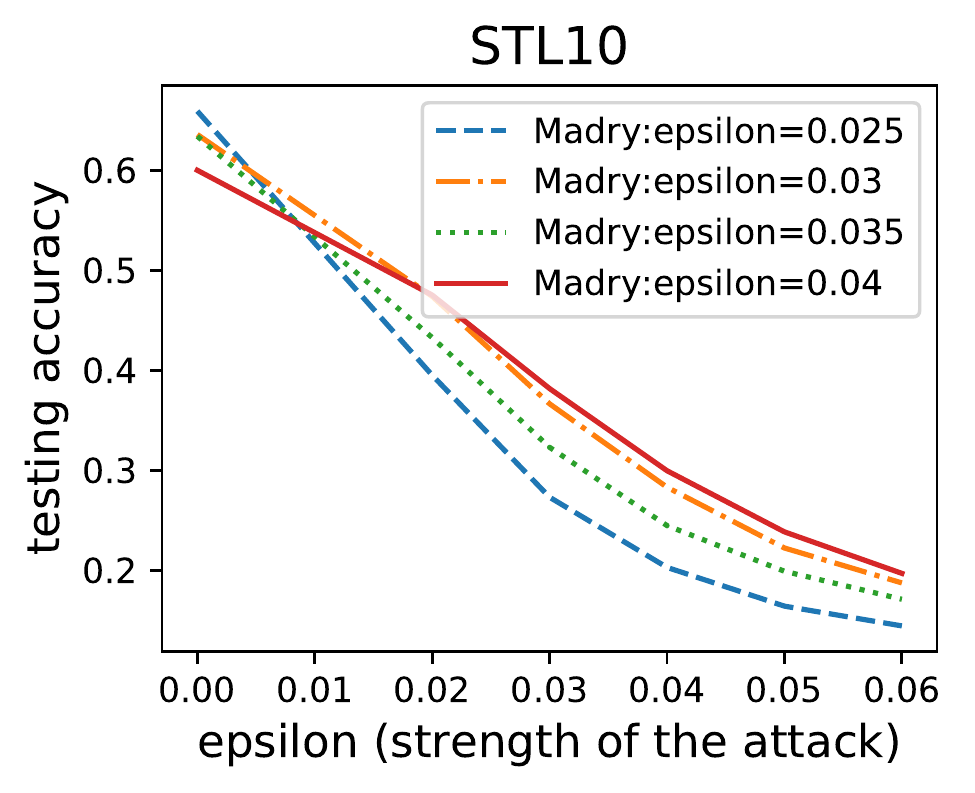}
    \caption{Testing accuracy of models with different $\epsilon$ on MNIST, CIFAR10 and STL10.}
    \label{fig:epsilon}
\vspace{15pt}
\end{figure}

Based on Figure~\ref{fig:epsilon}, we use $\epsilon=0.3, 0.03, 0.03$ in Madry's adversarial training on MNIST, CIFAR10 and STL10 respectively. For Tiny Imagenet, we use $\epsilon=0.01$. To make a fair comparison, we use the same $\epsilon$ when training OT-CLA+Adv. Code for OT-Classifier will be made public later on github.

\vspace{15pt}
\subsection*{B. Embedding Visualization}
\vspace{15pt}
In this subsection, we show larger version of Figure~\ref{fig:embed} for clearer view. Figure~\ref{fig:mnist_ecla} and Figure~\ref{fig:mnist_ot} are embedding visualization plots on MNIST. Figure~\ref{fig:cifar_ecla} and Figure~\ref{fig:cifar_ot} are embedding visualization plots on CIFAR10. The plot on the left is the 2D embedding generated from the legitimate images. The one on the right is the 2D embedding generated from adversarial images. Same for all four plots.

\begin{figure}
    \centering
    \includegraphics[width=0.45\textwidth]{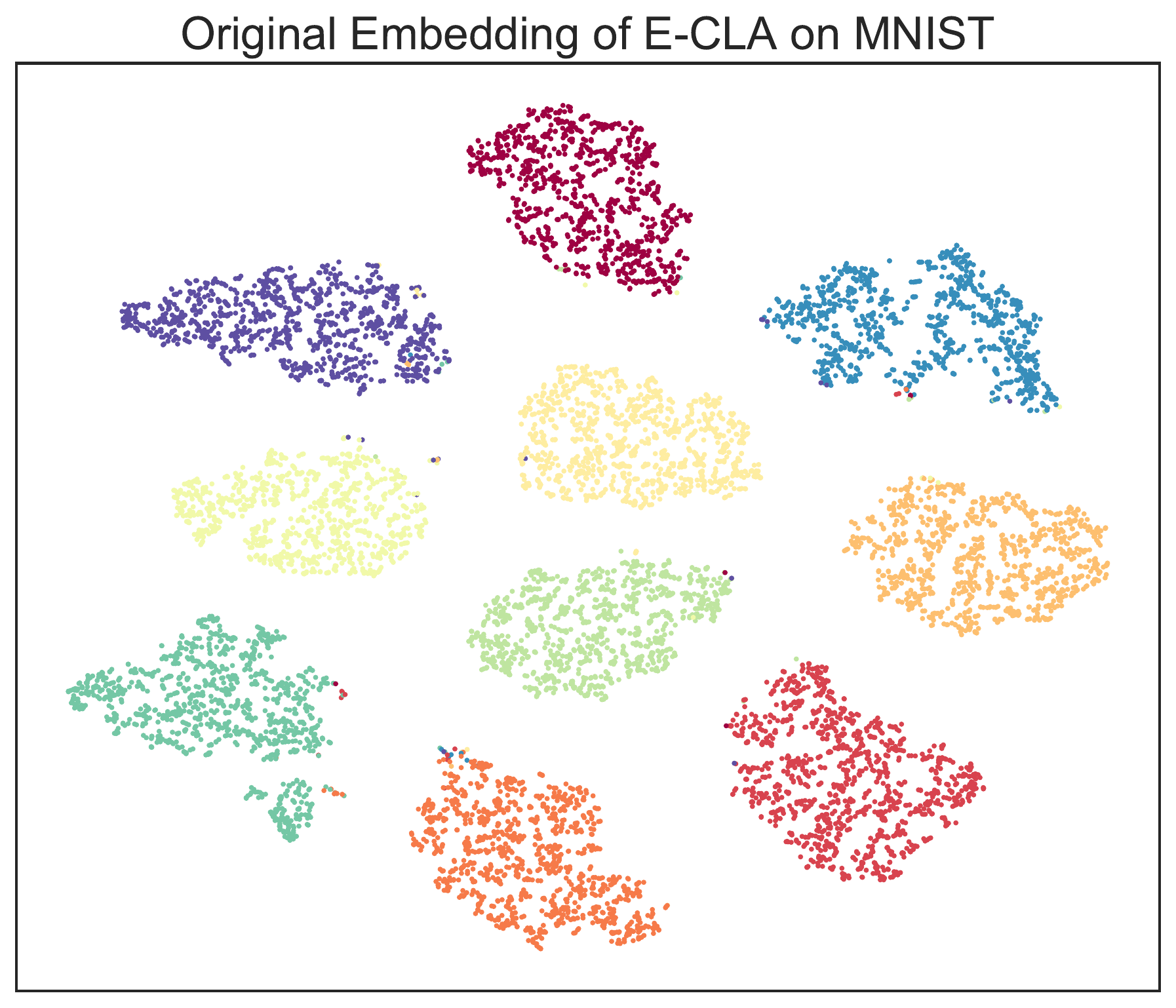}
    \includegraphics[width=0.45\textwidth]{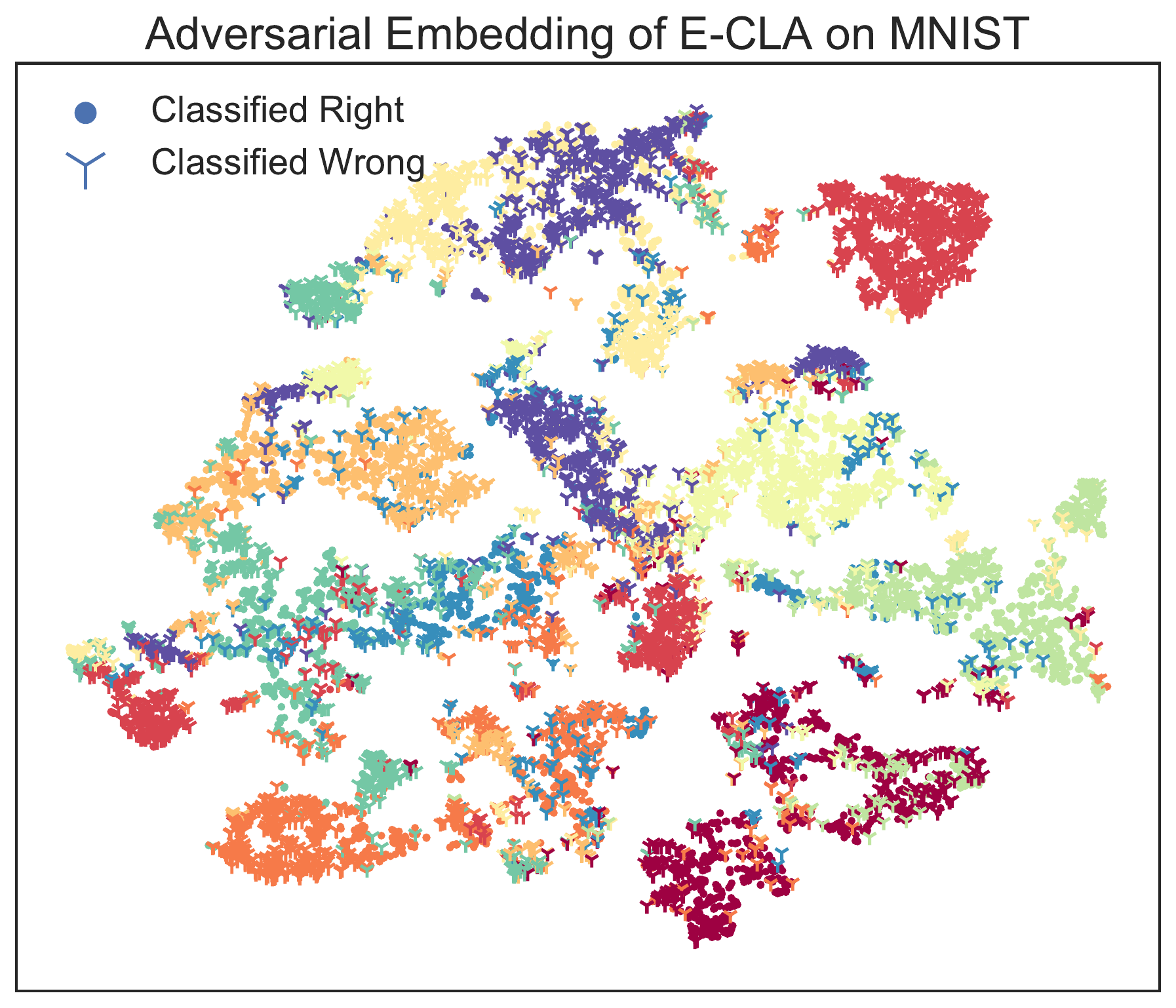}
    \caption{2D embedding for E-CLA on MNIST. }
    \label{fig:mnist_ecla}
\end{figure}

\begin{figure}
    \centering
    \includegraphics[width=0.45\textwidth]{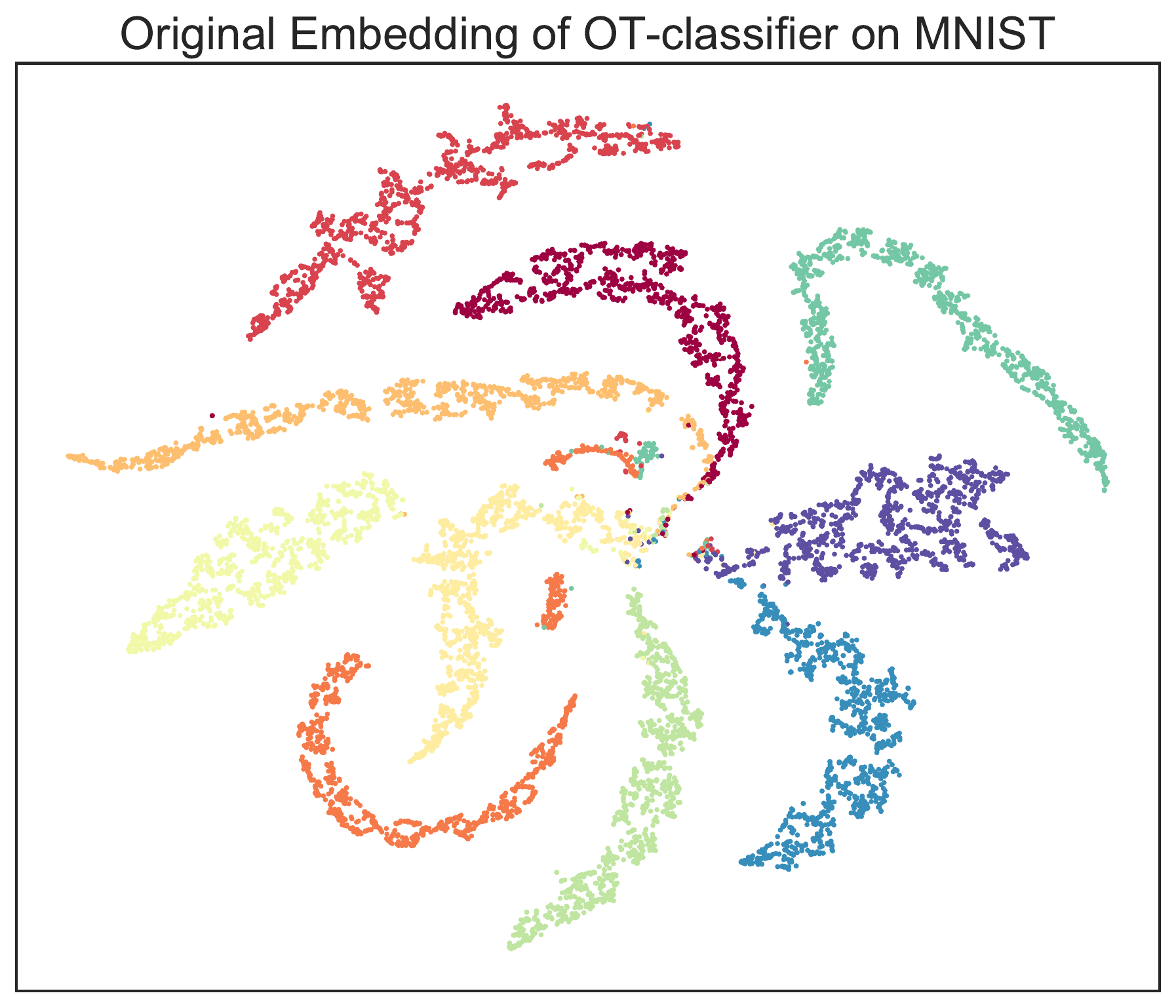}
    \includegraphics[width=0.45\textwidth]{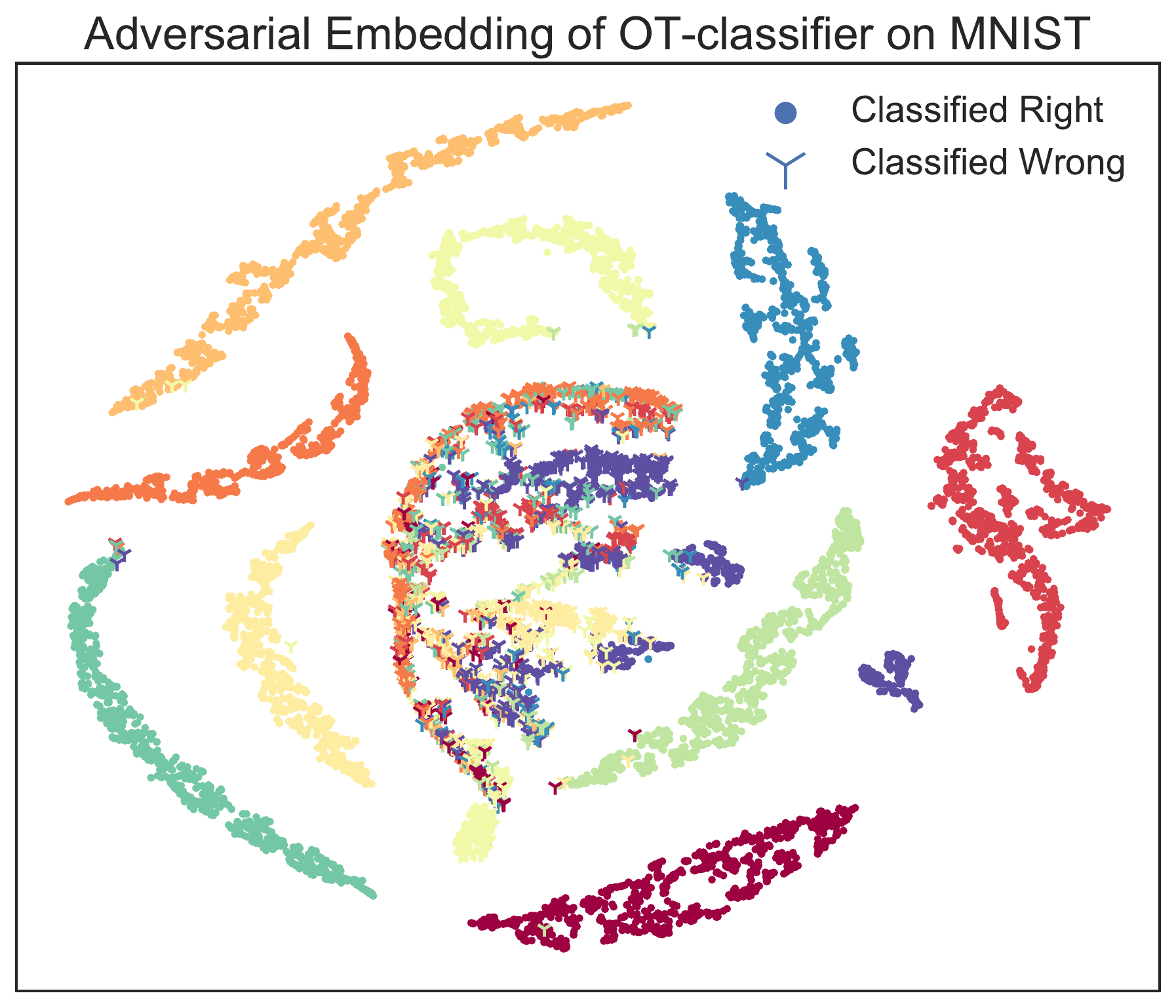}
    \caption{2D embedding for OT-Classifier on MNIST.}
    \label{fig:mnist_ot}
\end{figure}

\begin{figure}
    \centering
    \includegraphics[width=0.45\textwidth]{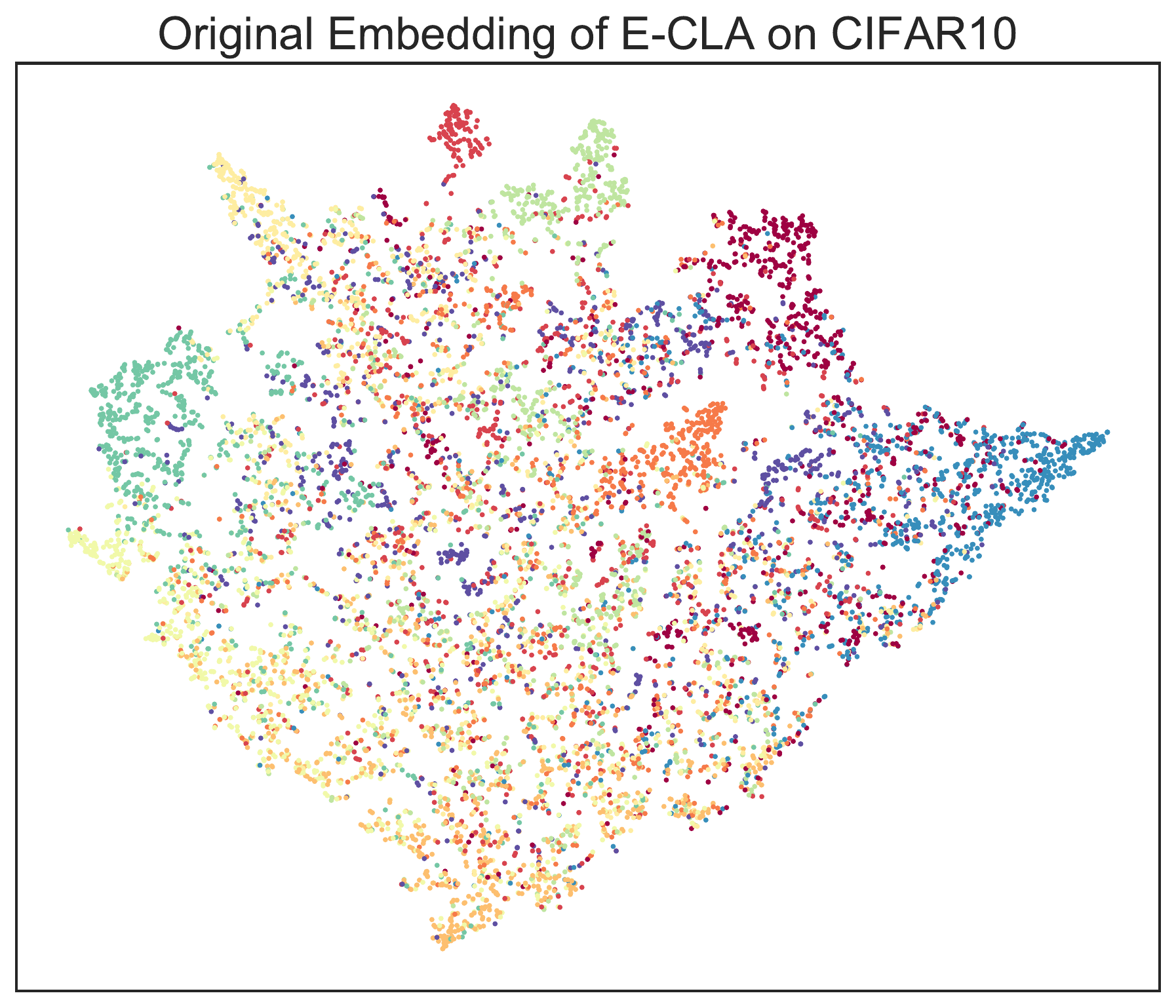}
    \includegraphics[width=0.45\textwidth]{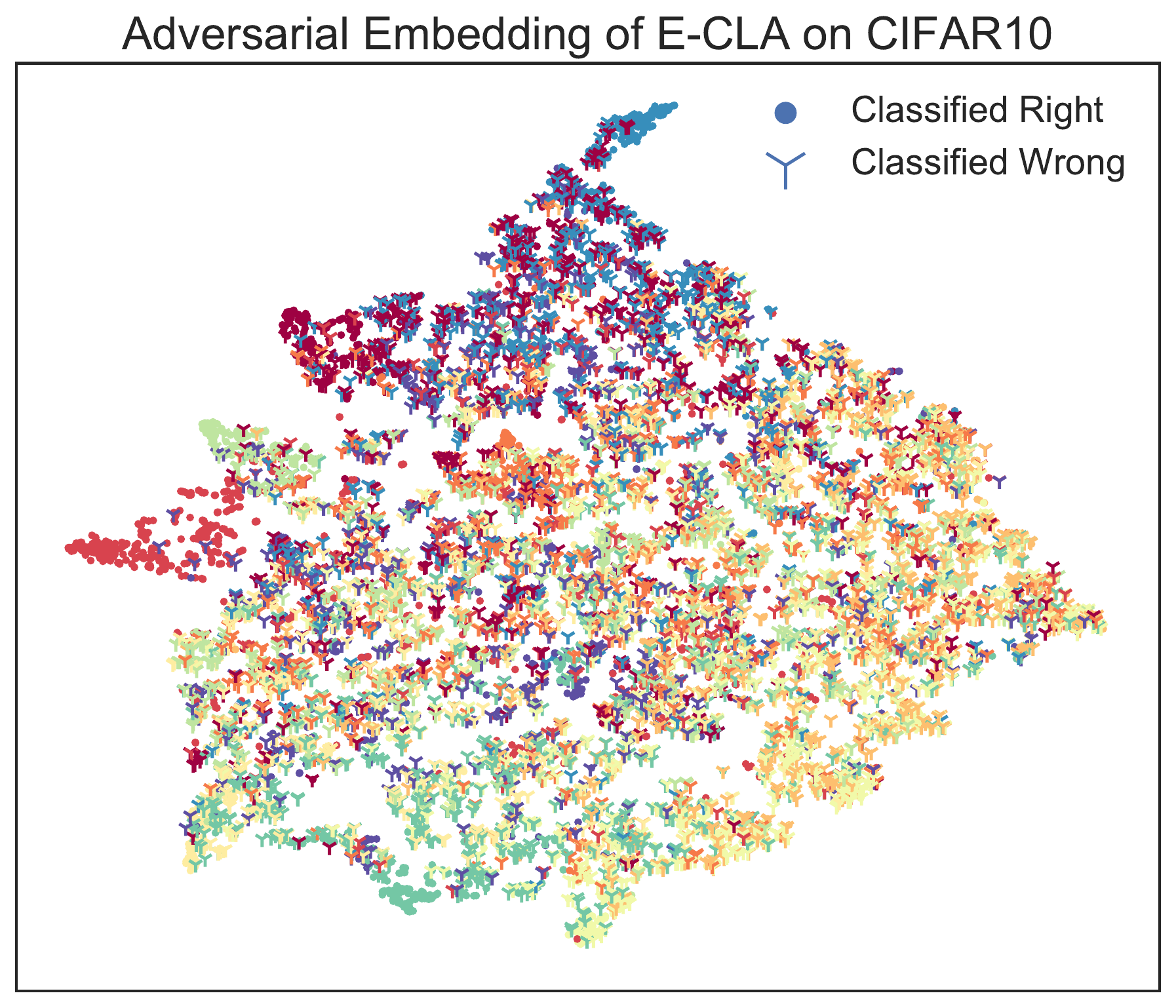}
    \caption{2D embedding for E-CLA on CIFAR10.}
    \label{fig:cifar_ecla}
\end{figure}

\begin{figure}
    \centering
    \includegraphics[width=0.45\textwidth]{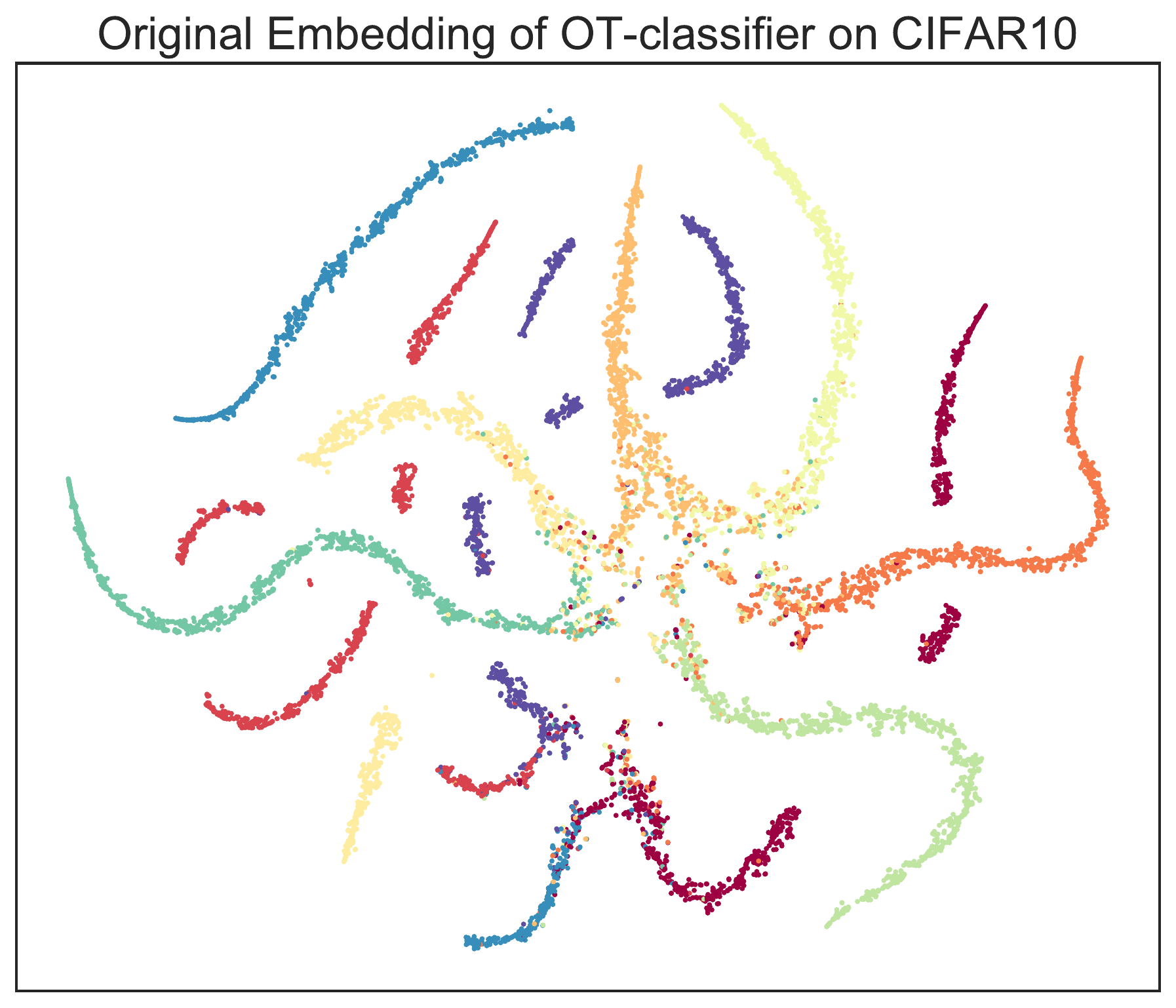}
    \includegraphics[width=0.45\textwidth]{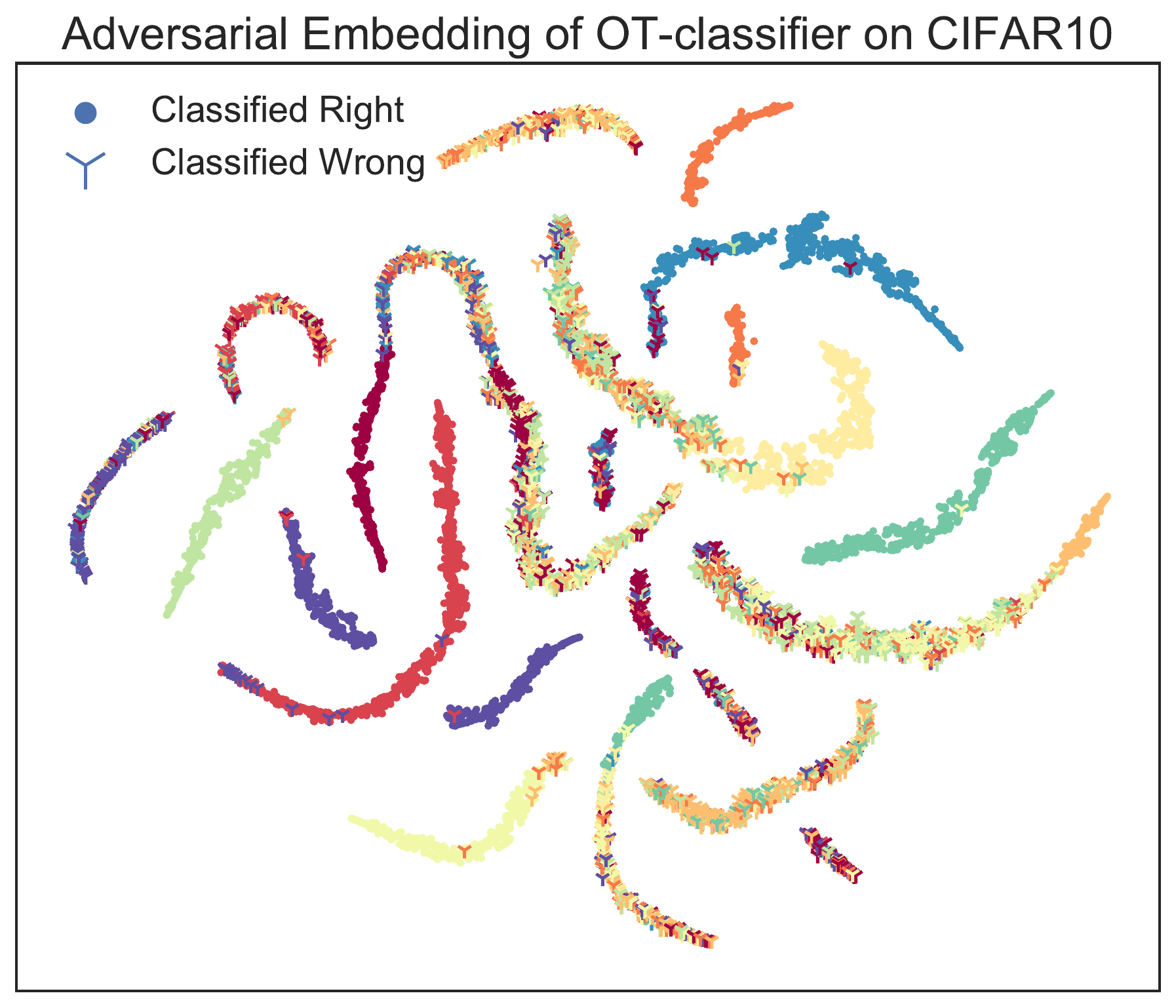}
    \caption{2D embedding for OT-Classifier on CIFAR10.}
    \label{fig:cifar_ot}
\end{figure}
    \endgroup
\end{document}